\documentclass[journal]{IEEEtran}
\usepackage{amsmath,amsfonts,amsthm}
\usepackage{algorithmic}
\usepackage{algorithm}
\usepackage{array}
\usepackage[caption=false,font=normalsize,labelfont=sf,textfont=sf]{subfig}
\usepackage{textcomp}
\usepackage{stfloats}
\usepackage{url}
\usepackage{verbatim}
\usepackage{graphicx}
\usepackage{cite}
\usepackage{multirow}

\usepackage{color}
\begin{document}

\title{RegHEC: Hand-Eye Calibration via Simultaneous Multi-view Point Clouds Registration \\ of Arbitrary Object}

\author{Shiyu Xing, Fengshui Jing, Min Tan
        % <-this % stops a space
\thanks{This work was supported by the National Natural Science Foundation of China under Grant No.62173327. Corresponding Author: Fengshui Jing}
\thanks{The authors are with School of Artificial Intelligence, University of Chinese Academy of Sciences, Beijing 100049, China, and also with the State Key Laboratory of Multimodal Artificial Intelligence Systems, Institute of Automation, Chinese Academy of Sciences, Beijing 100190, China (e-mail: xingshiyu2020@ia.ac.cn; fengshui.jing@ia.ac.cn; min.tan@ia.ac.cn).}}

\markboth{Journal of \LaTeX\ Class Files,~Vol.~14, No.~8, August~2021}%
{Shell \MakeLowercase{\textit{et al.}}: A Sample Article Using IEEEtran.cls for IEEE Journals}


\maketitle

\begin{abstract}
RegHEC is a registration-based hand-eye calibration technique with no need for accurate calibration rig but arbitrary available objects, applicable for both eye-in-hand and eye-to-hand cases. It tries to find the hand-eye relation which brings multi-view point
clouds of arbitrary scene into simultaneous registration under a common reference frame. RegHEC first achieves initial alignment of multi-view point clouds via Bayesian optimization, where registration problem is modeled as a Gaussian process over hand-eye relation and the covariance function is modified to be compatible with distance metric in 3-D motion space SE(3), then passes the initial guess of hand-eye relation to an Anderson Accelerated ICP variant for later fine registration and accurate calibration. RegHEC has little requirement on calibration object, it is applicable with sphere, cone, cylinder and even simple plane, which can be quite challenging for correct point cloud registration and sensor motion estimation using existing methods. While suitable for most 3-D vision guided tasks, RegHEC is especially favorable for robotic 3-D reconstruction, as calibration and multi-view point clouds registration of reconstruction target are unified into a single process. Our technique is verified with extensive experiments using varieties of arbitrary objects and real hand-eye system. We release an open-source C++ implementation of RegHEC.

\end{abstract}


\section{Introduction}
\IEEEPARstart{T}{he} recent advance of 3-D scanning technology has brought great potential for 3-D modeling applications, e.g. cultural heritage digitization\cite{Culturalheritages1,Culturalheritage2}, quality control\cite{QualityControl1,QualityControl2} and reverse engineering\cite{ReverseEngineering1,ReverseEngineering2}, to name a few.  Regardless of the final modeling purpose, scans must be taken from different viewpoints then aligned into a common coordinate system\cite{TRONBV} and the viewpoints selection is a key factor for the final modeling quality and completeness\cite{Autonomousmetrology}.
\begin{figure}[!t]
	\centering
	\includegraphics[width=0.5\textwidth]{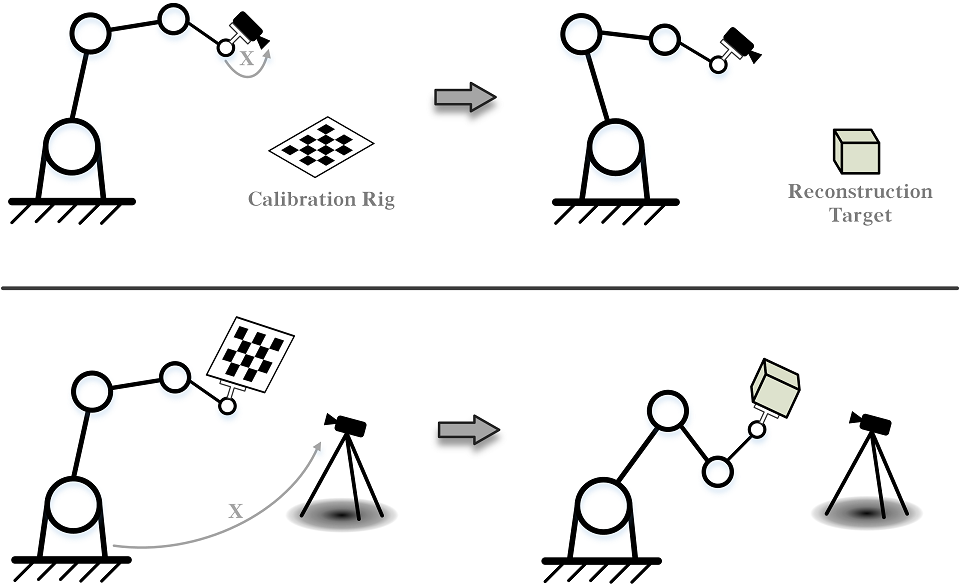}
	\caption{Conventionally, hand-eye calibration is performed as a separate process before the following reconstruction task, owing to the reliance on accurate calibration rig. Figure above and below show the eye-in-hand and eye-to-hand scenario respectively.}
	\label{fig_1}
\end{figure}

The traditional approach is largely manual and hand guided, where an experienced operator plans the next scanning view repeatedly according to a real-time visualization of the partially reconstructed model then carry out the scanning accordingly\cite{bodenmueller2009streaming}. However, empirically selecting views is not only tedious and time consuming, but even undesirable or impossible when facing geometrically and topologically complex shapes\cite{scott2003viewplanning}. 

To address this, active 3-D reconstruction, a typical use case of robotic hand-eye system consisting of 3-D sensor and serial manipulator, has been widely studied over decades. While most of the relevant research uses an eye-in-hand setup\cite{monica2018eyeinhand2,wang2019eyeinhand3,kriegeleyeinhand1} where 3-D sensor is fixed on the robot flange (the mounting plate where usually tool 0 frame is defined in default) and target to be modeled is placed within the workspace of the robot, experiments using object in hand and stationary 3-D sensor were also conducted\cite{wu2014eye2hand1,kobayashi2022obtainingEye2hand2,krainin2011Eye2hand3}. 

For both setup scenarios, next best view(NBV) calculation and point cloud registration are carried out alternately and iteratively as two critical processes until the reconstruction is complete\cite{kriegel2015NBV}. To control the robot configuration according to the calculated NBV and achieve alignment of captured point cloud to the partially complete model, the flange-sensor relation in eye-in-hand case or base-sensor relation in eye-to-hand case need to be accurately recovered, which is generally called hand–eye calibration.

Even though the sensor is firmly mounted on the robot flange or fixed next to the robot, the hand-eye transformation may change in some occasions\cite{peters2020extrinsic}. Thus, hand-eye calibration, which seems once in-a-lifetime task, needs to be conducted on-site from time to time. Commonly used approaches rely on either accurate calibration rig with known dimensions or special geometrical constraints. In that case, hand-eye calibration must be performed beforehand as a separate process\cite{Peng,Tang}, leading to cost and time pressure, as shown in Fig.1.  

We propose RegHEC, a registration-based hand-eye calibration technique using multi-view point clouds of arbitrary object with no prior information, applicable for both eye-in-hand and eye-to-hand scenarios. It uses reconstruction target directly as calibration object, achieving both calibration and multi-view point clouds registration simultaneously, thus shaking off the unnecessary separation. RegHEC is applicable for most 3-D vision guided tasks with special superiority for robotic 3-D reconstruction, and is even feasible for object without distinctive 3-D features like sphere, cone, cylinder and simple plane, which can be quite challenging for correction point cloud registration and sensor motion estimation using pair-wise ICP algorithm.

The contributions of this paper are four-fold:
\begin{enumerate}
	\item{An Anderson accelerated ICP variant(AA-ICPv) was put forward for faster multi-view point clouds registration and hand-eye calibration. We show that our recently proposed ICP variant\cite{xing2022TII}, which aligns multi-view point clouds into robot base frame by iteratively refining hand-eye relation, is essentially a fixed-point problem and significantly speed up its convergence applying Anderson Acceleration. AA-ICPv does not only use the last iterate but tries to predict the most plausible convergence point based on the history of previous iterates and their residuals.}
	\item{A Bayesian optimization based initial alignment (BO-IA) method was proposed to give a proper initial guess of the hand-eye relation for the later ICP variant with Anderson Acceleration. We modeled the registration error over hand-eye relation as a Gaussian Process  and modified the covariance function to render it compatible with distance metric in 3-D motion space SE(3), making it more efficient searching the global optimal with acquisition function.}
	\item{By controlling the robot to present different angles of the object in the hand to the stationary sensor and switching the registration reference frame to robot flange frame, we successfully extend the use case into eye-to-hand scenario, making it a complete and general solution for 3-D sensor. Besides, when observing empty hand(taking the end-effector as reconstruction/calibration object), the tool center point(TCP) position can be obtained in the meanwhile.}
	\item{We release an open-source C++ implementation of resulting algorithm named RegHEC\footnote{Source code is freely available at https://github.com/Shiyu-Xing/RegHEC}. Extensive experiments with real hand-eye system verify our technique feasible and effective for both eye-in-hand and eye-to-hand setup. We showcase that RegHEC is applicable even facing solid of revolution (SOR) like cylinder, cone and sphere, or very basic geometry like simple plane. }
\end{enumerate}

\section{Related Work}
As a classic problem in robotics and mathematics, hand-eye calibration has been studied extensively for decades. It was first proposed by Shiu\cite{Shiu} in 1980s and formulated as the widely-known matrix equation AX=XB, where A and B are homogeneous transformation matrices that relate robot hand and sensor at different time frames respectively, and X is the hand-eye relation to be recovered.

At an early stage, the transformation matrix equation was decomposed into a purely rotational and translational part then solved separately via unit quaternions\cite{Chou}, rotation axis and angle\cite{Tsai}, and Lie Theory\cite{Park}. These research studies all claimed that the problem is solvable with linear algebra, provided at least two robot motions with unparallel rotation axes. However, these separate solution inevitably causes error propagation from rotation estimation to translation, which led to the later development of simultaneous solution treating the relative position and orientation in a unified way, in both analytical form\cite{DQsimulClose,KroneckerSimulClose,chen1991screwSimulClose} and iterative form\cite{ShiuSimulIterative,HoraudSimulIterative,convexZhaoSimulIterative}.

The essence of solution family based on AX=XB is kinematic loop formed by serval rigid body motions, which is applicable for both eye-in-hand and eye-to-hand cases regardless of sensor type and its working principle. It is still the most commonly used hand-eye calibration technique due to its versality and robustness. While the experiments in above mentioned researches are designed with accurate calibration rig for camera motion estimation, AX=XB also finds application in scenarios with 3-D sensors and unstructured environment. An attempt given by Xu\cite{SceneFeatures} applied Perspective-n-Point algorithm and bundle adjustment using scene features to estimate of motions of the RGB-D camera, then solved AX=XB. Although this method was verified feasible from application perspective via a robotic grasping experiment, it is not applicable for 3-D sensor without image output and fails to cope with eye-to-hand setup and texture-less scenario which can be quite common for 3-D reconstruction task. Wenberg\cite{wenberglidarlidar} applied pair-wise point cloud registration to recover sensor motion then successfully calibrated the lidar-lidar extrinsic parameters by solving an AX=XB based optimization problem. While the idea can be easily transferred to robotic hand-eye scenario\cite{PatentCN}, it finds limitation when calibration object is symmetric or SOR like cylinder, cone and sphere, with which point clouds registration and sensor motion estimation using existing pair-wise ICP can be far from accurate. 

Another formulation using absolute transformations, first proposed by Zhuang\cite{zhuang1994AXYBfirst}, can be expressed as AX=YB, where A and B no longer specify the relative motion between different time instances but transformation from hand to robot base and sensor to world frame respectively. This branch relies on accurate calibration rig to define the world frame, and robot-world transformation Y is to be determined in addition to hand-eye relation X as byproduct, thus such formulation is widely known as simultaneous hand-eye and robot-world calibration.
Similarly, both separate solution\cite{HoraudAXYBIterative,shah2013AXYBseparate}, which decouples translation from rotation, and simultaneous solution\cite{li2010AXYBsimul,HoraudAXYBIterative,Weighted}, which integrates both rotation and translation estimation, were studied extensively.
Recently, some reprojection error based methods\cite{tabb2017AXYBreproj,koide2019RALreproj,pedrosa2021TROatomicReproj} were proposed to tackle AX=YB problem and were reported to present higher calibration accuracy compared with conventional methods, whereas a checkerboard to define the world frame and image output are still indispensable.

Hand-eye calibration techniques for 3-D sensor specifically were also put forward, whereas calibration rig with high manufacturing accuracy and known dimensions is a must in most of relevant literatures. Typical instances include calibration board\cite{TIEstichingChecker}, standard sphere\cite{sphere3} and free-form surface with design CAD model\cite{xie2021generalFreeformSurface,2D3D}. In addition to the costly high-precision machining, these methods can be unapplicable due to unavailability and accuracy degeneration of calibration rig. Method using common geometry like plane and its normal vector to make calibration rig dispensable was reported in\cite{Plane1}. Nevertheless, auxiliary tool is needed to measure points coordinates on the plane beforehand, which brings extra operational complexity. Limoyo\cite{ContactMap} proposed a eye-to-hand self-calibration method,
which estimates robot-sensor relation by aligning a contact-based point cloud generated by manipulator and vision-based point cloud generated by the sensor. Although experiments were conducted successfuly with two rectangular boxes and on-board hardware only, the contact-based point cloud generation using manipulator under impedance control is rather time consuming, and this method can not be applied facing SOR and simple plane, as the registration is realized by pair-wise ICP.

To the best of our knowledge, RegHEC is the first 3-D sensor hand-eye calibration technique that 1) uses arbitrary available object with little requirement(can be SOR or plane), 2) relies on 3-D measurements only, and 3) works in both eye-in-hand and eye-to-hand cases. It returns the calibrated hand–eye relation together with multi-view point clouds registration, thus is more beneficial for robotic 3-D reconstruction than for other 3-D vision-guided tasks.

\section{Methodology}
Our idea is based on the fact that point clouds captured at different viewpoints align, after transformation into robot base frame by left multiplying first correct hand-eye(flange-sensor) relation then corresponding robot poses(pose of flange frame w.r.t robot base, when point clouds are captured), as is shown in Fig.2. Better hand-eye estimation means better registration and vice versa. Therefore, we reformulate the hand eye calibration of 3-D sensor as finding the hand-eye relation, which simultaneously register multi-view point clouds of stationary scene into a common reference frame, using corresponding robot poses where point clouds were captured. 

In the following subsections, we first briefly recap an important least square problem for essential background. Then Anderson accelerated ICP variant, initial guess estimation using Bayesian optimization and extension to eye-to-hand case are introduced in sequence. Finally, some implementation details are given.

\begin{figure}[!t]
	\centering
	\includegraphics[width=0.44\textwidth]{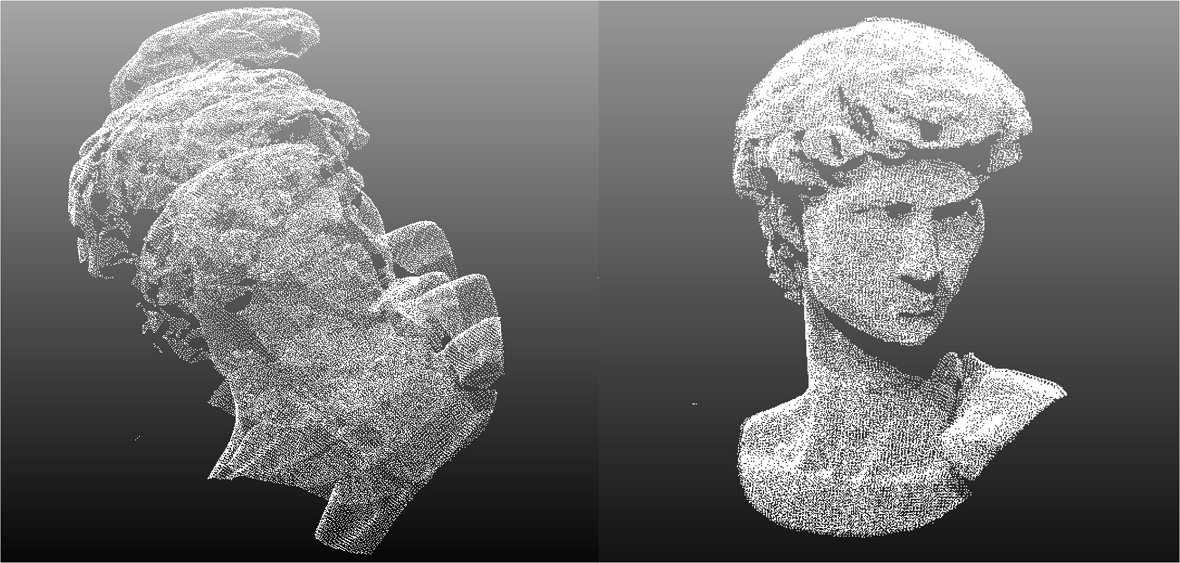}
	\caption{Point clouds of a David plaster figure were captured by a eye-in-hand system, from 5 different viewpoints, then transformed into the robot base frame with hand-eye relation and corresponding robot poses where point clouds were captured. Incorrect hand-eye relation presents obvious misalignment after transformation (left) while correct hand-eye relation showcases fine registration(right.)}
	\label{fig_3}
\end{figure}

\subsection{Least square alignment of multi-view point sets with known correspondences}
We briefly recap this least square problem formualted in our previous work\cite{xing2022TII}, as it provides indispensable backgroud.

Consider a serial manipulator with a rigidly mounted 3-D sensor making $n$ motions from $1$st robot pose $\boldsymbol A_1$ to $(n+1)${th} robot pose $\boldsymbol A_{n+1}$ (that is, first motion moves from the first robot pose to the second, second motion is from the second pose to the third, and so on). The robot pauses at each pose to capture a point cloud of the stationary scene. We then pair the point clouds captured before and after each robot motion and match the corresponding points in each pair. 

Fig.3 shows an example of $i$th robot motion from $i$th pose to $(i+1)$th pose, with $\{({\boldsymbol p_j^i},{\boldsymbol q_j^i})\}$ denoting the set of correct correspondences that we assume is known. $({\boldsymbol p_j^i},{\boldsymbol q_j^i})$ stands for the $j$th correspondence in $i$th robot motion, where $\boldsymbol p_j^i$ and $\boldsymbol q_j^i$ are coordinates of a common scene point under sensor frame at $i$th robot pose (before $i$th motion) and $(i+1)$th robot pose (after $i$ th motion) respectively. 
\begin{figure}[!t]
	\centering
	\includegraphics[width=0.44\textwidth]{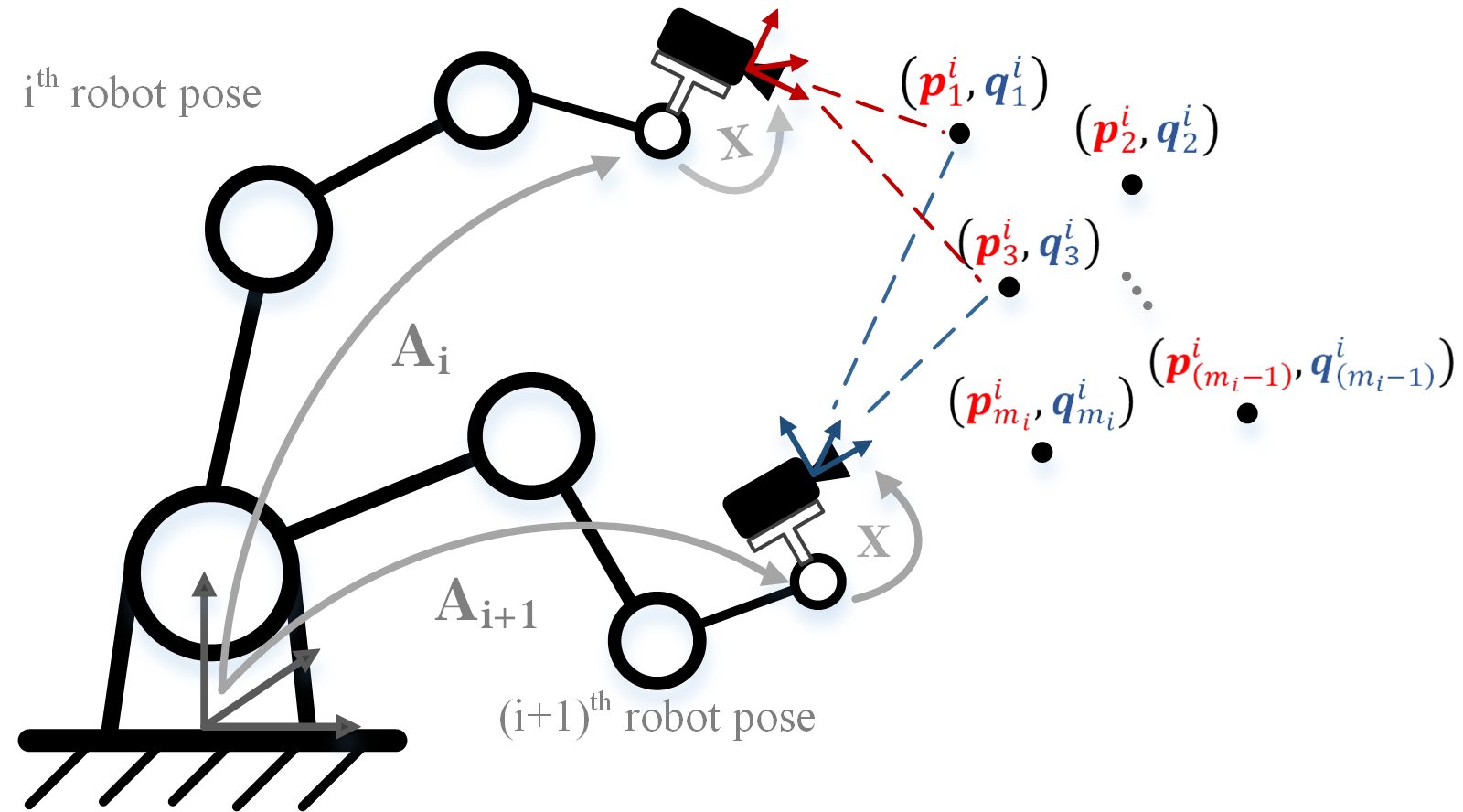}
	\caption{The $i$th motion of eye-in-hand system moves from $i$th robot pose to $(i+1)$th robot pose and $m_i$ 3D-3D correspondences are established between point clouds captured at this two robot poses. $\boldsymbol p_1^i$, $\boldsymbol p_2^i$...$\boldsymbol p_{m_i}^i$ (in red) represent the measured 3-D coordinates of scene points w.r.t sensor frame(in red) at $i$th pose and  $\boldsymbol q_1^i$, $\boldsymbol q_2^i$...$\boldsymbol q_{m_i}^i$ (in blue) denote corresponding measured 3-D coordinates w.r.t sensor frame(in blue) at $(i+1)$th pose.}
	\label{fig_3}
\end{figure}

Ideally, the hand-eye relation, represented by a homogeneous transformation matrix $\boldsymbol X$ in SE(3), aligns the 3D-3D correspondences obtained from any camera motion, under robot base frame. Thus, considering all the correspondences in all $n$ pairs of point clouds (formed by $(n+1)$ point clouds captured during $n$ robot motions). Hand eye relation can be estimated by minimizing the sum of squared-error on the distance between corresponding points after transformation into robot base frame, in a least square sense. In standard form:
\begin{equation}
			\begin{split}
				\begin{aligned}	
					&\!\!min\,\,\,\, f(\boldsymbol R_X ,\boldsymbol t_X)=\sum_{i=1}^n\sum_{j=1}^{m_i}{||g_{ij}(\boldsymbol R_X, \boldsymbol t_X)||}^2\\
					&\!s.t. \quad\,\boldsymbol R_X \in \text{SO(3)} ,\quad \boldsymbol t_X \in \mathbb{R}^3
				\end{aligned}
			\end{split}
\end{equation}
in which
\begin{equation}
	\\[1ex]	
	{\begin{small}	
			\begin{split}	
				\begin{aligned}
					\!\!&\boldsymbol g_{ij}(\boldsymbol R_X \!,\!\boldsymbol t_X)\\[1ex]
					\!\!&= 
					\begin{bmatrix}
						% 	\begin{smallmatrix}
						\boldsymbol R_{A_i}\!\!& \!\!\boldsymbol t_{A_i}\\ 
						%	\end{smallmatrix}
					\end{bmatrix}
					\begin{bmatrix}
						%	\begin{smallmatrix}
						\boldsymbol R_{X}\!\!& \!\!\boldsymbol t_{X}\\ 
						\boldsymbol 0\!\!&\!\! 1 
						%	\end{smallmatrix}
					\end{bmatrix}
					\begin{bmatrix}
						% 	\begin{smallmatrix}
						\boldsymbol p_j^i\\ 
						1 
						%	\end{smallmatrix}
					\end{bmatrix}
					\!-\!\begin{bmatrix}
						%	\begin{smallmatrix}
						\boldsymbol R_{A_{i+1}}\!\!&\!\! \boldsymbol t_{A_{i+1}}\\ 
						%	\end{smallmatrix}
					\end{bmatrix}
					\begin{bmatrix}
						%	\begin{smallmatrix}
						\boldsymbol R_{X}\!\!&\!\! \boldsymbol t_{X}\\ 
						\boldsymbol 0\!\!&\!\! 1 
						%	\end{smallmatrix}
					\end{bmatrix}
					\begin{bmatrix}
						% 	\begin{smallmatrix}
						\boldsymbol q_j^i\\ 
						1 
						%	\end{smallmatrix}
					\end{bmatrix}\\[1ex]
					\!\!&=\boldsymbol R_{A_i}\boldsymbol R_{X}\boldsymbol p_j^i \!+\!\boldsymbol R_{A_i}\boldsymbol t_{X}\!+\!\boldsymbol t_{A_i}\!-\!\boldsymbol R_{A_{i+1}}\boldsymbol R_{X}\boldsymbol q_j^i \!-\!\boldsymbol R_{A_{i+1}}\boldsymbol t_{X}\!-\!\boldsymbol t_{A_{i+1}}
				\end{aligned}
			\end{split}	
	\end{small}}
\end{equation} 
The above $n$ is the number of robot motions, i.e. the number of point cloud pairs in which we establish correspondences, and $m_i$ denotes the number of correspondences in $i$th motion or say $i$th point cloud pair. A thorough mathematical derivation of perturbation-based Gauss-Newton method for solving this least square problem is given in Appendix.

\subsection{An Anderson accelerated ICP variant(AA-ICPv) for multi-view point clouds registration} 
Our goal is to manage more general case in which correct 3D-3D correspondences in multi-view point clouds can be hardly obtained beforehand and the above least square technique can not be applied as a standalone hand-eye calibration method. Such cases can be texture-less scenario or using 3-D sensor without image output.
\begin{figure*}[!t]
	\centering
	\includegraphics[width=1\textwidth]{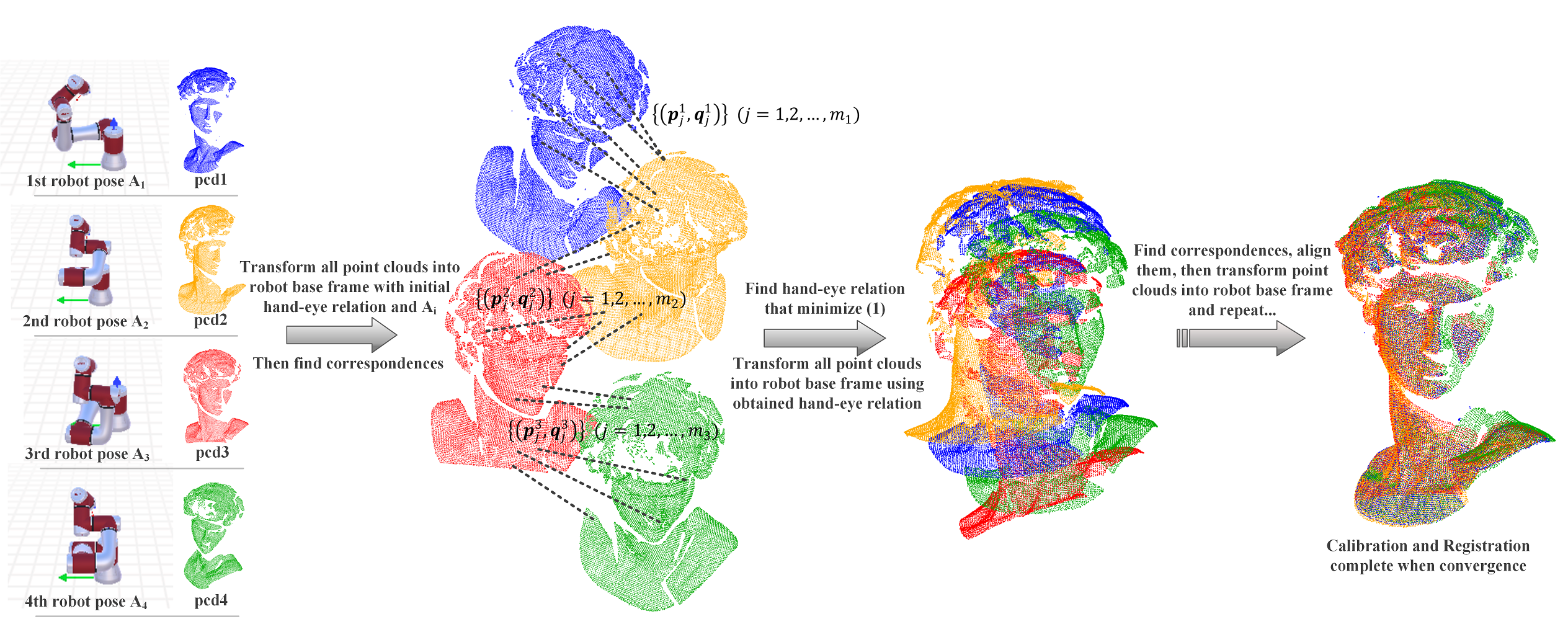}
	\caption{An example of proposed ICP variant in the case of point clouds captured at 4 different robot poses. Transform all point clouds into robot base frame, establish correspondences($\{({\boldsymbol p_j^1},{\boldsymbol q_j^1})\}(j=1,2,...,m_1)$ denotes the set of $m_1$ correspondences between pcd1 and pcd2, with $\boldsymbol p_j^1$ indicating point in pcd1 and $\boldsymbol q_j^1$ its corresponding point in pcd2, and so on), then find hand-eye relation that minimize distance between corresponding points, i.e. solve (1). These processes are repeated until convergence. It returns the calibrated hand-eye relation and registration of 4 point clouds simultaneously.}
	\label{fig_4}
\end{figure*}

We find inspiration in research on pair-wise registration(pose estimation, alignment, motion estimation) of 3-D free-from shapes, especially the iterative closest point (ICP) algorithm which 
registers two point clouds with unknown correspondences by iteratively alternating between correspondences estimation and alignment\cite{horn1987pointsetsQuaternion,SVD2pointsets}.
The most classic and well known ICP algorithm\cite{ICPBesl} can be summarized as follows:
\begin{enumerate}
	\item{For given floating source point cloud S and fixed reference point cloud R, transform S into the coordinate frame of R using initial transformation estimate $\boldsymbol T^0$.}
	\item{Correspondences step: For each point in S find the closest corresponding point in R,  pairs of such points are called correspondences.}
	\item{Alignment step: Find such transform $\boldsymbol T$ that minimizes the mean of squared distance between correspondences (i.e. mean square error(MSE)).}
	\item{Transform the source point cloud S using the obtained transformation.}
	\item{Go to step 2) and repeat until convergence.}
\end{enumerate}
Although various modifications have been made, they are usually either implemented in the correspondences step\cite{chetverikov2005TRICP1,serafin2015nicp,rusinkiewicz2001efficientICP} or alignment step\cite{chen1992Point2planeICP1,bouaziz2013sparseICP,Low2004LinearPoint2Plane}, and the above framework still holds, which alternates between corresponding point query in step 2) and objective minimization in step 3). 

Given our motivation of finding hand-eye relation which registers multiple point clouds of a common scene captured from different viewpoints into robot base frame and inspired by the ICP framework, we proposed an ICP variant which iteratively estimate correspondences based on closest point criteria in multi-view point clouds then refines hand-eye relation using method in previous subsection to bring the correspondences into alignment.
This algorithm is summarized as follows:
\begin{enumerate}
	\item{Consider the hand-eye system in previous subsection again which makes $n$ robot motions from $1$st robot pose to $(n+1)$th robot pose and pauses at each robot pose to capture a point cloud.}
	\item{Transform all $(n \!+\!1)$ point clouds into the robot base frame by left multiplying first the initial guess of hand-eye relation $({\boldsymbol R_\text X}^0,{\boldsymbol t_\text X}^0)$ then corresponding robot pose $\boldsymbol A_i$.}
	\item{Correspondences step: Pair point clouds captured before and after each robot motion. In each pair, for each point in the smaller point cloud(point cloud with less points), we find its corresponding closest point in the other point cloud to form one correspondence (${\boldsymbol p_j^i},{\boldsymbol q_j^i}$).}
	\item{Alignment step: Find such hand-eye transformation which minimizes the mean of squared distance between corresponding points, i.e. minimize MSE by solving problem (1) in previous subsection A.}
	\item{Transform all $(n \!+\!1)$ point clouds into robot base frame using the obtained hand-eye relation and corresponding robot poses.}
	\item{Go to step 3) and repeat until convergence}
\end{enumerate}
 
Fig.4 illustrates the ICP variant in the case of 4 point clouds captured. While inspired by the genius idea of classic pair-wise ICP that registration can be done by iteratively estimating correspondences then solving optimization problem to bring them into alignment, the ICP variant is still very different in details. The classic pair-wise ICP aligns two point clouds, under the coordinate frame of reference cloud, by iteratively refining the sensor motion/object pose to be estimated. The proposed ICP variant aligns multi-view point clouds simultaneously, under robot base frame, by iteratively refining hand-eye relation. Once the
registration is done, we believe the hand-eye relation is also calibrated at the same time.
Some degenerative cases include single robot motion(two robot poses), pure translation motion and multiple motions with parallel rotation axes\cite{xing2022TII}.

Mathematically, this ICP variant can be described as a mapping iteration of hand-eye relation variables $\boldsymbol R_X$ and $\boldsymbol t_X$
\begin{equation}
	({\boldsymbol R_\text X}^{k+1},{\boldsymbol t_\text X}^{k+1})= G(({\boldsymbol R_\text X}^{k},{\boldsymbol t_\text X}^{k}))	
\end{equation}
where $G$ call transforms all point clouds to robot base frame using $({\boldsymbol R_\text X}^{k},{\boldsymbol t_\text X}^{k})$(step 5)), estimate correspondences(step 3)) then returns the next hand eye relation $({\boldsymbol R_\text X}^{k+1},{\boldsymbol t_\text X}^{k+1})$ that minimize the distance between corresponding points(step 4)). 
The $G$ call is applied repeatedly until the difference between two adjacent iterates is small enough, spawning a Cauchy iteration sequence which approaches a fixed point $({\boldsymbol R_\text X}^*,{\boldsymbol t_\text X}^*)$. Therefore, the ICP variant essentially boils down to a fixed-point problem of finding $({\boldsymbol R_\text X}^*,{\boldsymbol t_\text X}^*)$ at which
\begin{equation}
	({\boldsymbol R_\text X}^*,{\boldsymbol t_\text X}^*)= G(({\boldsymbol R_\text X}^*,{\boldsymbol t_\text X}^*)).	
\end{equation}

This “state-less” approach, which calculates next iterate simply based on the last iterate, is usually low efficient especially when iterate is close to the fixed point, so our idea is to alleviate the slow convergence using Anderson Acceleration(AA)\cite{walkerNI2011anderson,anderson1965integraequations}, a well-established technique to speed up convergence of fixed-point problem. 

Given a fixed-point iteration $x^{k+1} \!=\! g(x^k)$, define the residual at $x^k$ as $f^k\!=\!g(x^k)-x^k$, which must vanish at fixed point $x^*$. Anderson acceleration does not only utilize the current iterate $x^k$, but also takes the previous $m$ iterates $x^{k-1},...,x^{k-m}$ into consideration to derive the next iterate $x^{k+1}$ which decrease the residual as much as possible. Specifically, the accelerated iterate is given as
\begin{equation}
		x^{k+1}=\sum_{i=0}^m\alpha_i g(x^{k-m+i})
\end{equation}
It lies on an affine subspace spanned by $g(x^{k-m}),...,g(x^{k})$ with $\boldsymbol \alpha =(\alpha_0,...,\alpha_m)$ being the affine coordinates, which is easily obtained by solving least square problem
\begin{equation}
	{
		\begin{split}
				\begin{aligned}	
					&\!\!min\,\,\,\,  ||\sum_{i=0}^m \alpha_i f^{k-m+i}||_2\\
					&\!\!s.t. \quad\, \sum_{i=0}^m \alpha_i=1
				\end{aligned}
		\end{split}
	}
\end{equation}
In practice, the equality constraint can be replaced by substitution $\alpha_m = 1-\sum_{i=0}^{m-1} \alpha_i $   which leads to the common unconstrained form
\begin{equation}
				\!\!min\,\,\,\,  ||f^{k}+\sum_{i=0}^{m-1} \alpha_i (f^{k-m+i}-f^{k})||_2\\
\end{equation}

Anderson acceleration attempts to predict the most plausible convergence point based on the history of previous iterates. It jumps across many unnecessary $g$ call, which is much more time-consuming compared with the negligible overhead of solving linear least square problem (7), thus convergence is much accelerated. This strategy is especially efficient when large number of point clouds are processed by the ICP variant as computational cost in each $G$ call can be heavier in that case.

Nevertheless, using $({\boldsymbol R_\text X}^{k},{\boldsymbol t_\text X}^{k})$ to parameterize rigid hand-eye transformation is invalid, as SO(3) is defined on multiplication and affine combination of multiple rotation matrices in general is not a rotation matrix itself. A parameterization using Euler angles was reported in\cite{pavlov2018AAICP}, but small rotation is assumed to avoid instability near singularity. Another possible parameterization is unit quaternion, which, like rotation matrix, however, is a Lie group defined on multiplication thus affine combination doesn’t result in unit vector in general. We notice that Lie algebra so(3), the tangent space at the identity transformation in SO(3), is a vector space itself, thus we can parameterize the hand-eye transformation during AA by concatenating the lie algebra of rotation and translation vector into 6-dimensional vector. That is 
\begin{equation}
\boldsymbol	u^k=\begin{bmatrix}
		log(\boldsymbol R^k)&\boldsymbol t^k
	    \end{bmatrix}=\begin{bmatrix}
	    \boldsymbol v^k&\boldsymbol t^k
    \end{bmatrix}
\end{equation}
where $log(\cdot)$ specifies the logarithm mapping from SO(3) to so(3)\cite{sola2018microLie}. To bridge over the parameterization in $\boldsymbol	u^{k}$ and $({\boldsymbol R_\text X}^{k},{\boldsymbol t_\text X}^{k})$, define $\widetilde{G}$ call as
\begin{equation}
	\boldsymbol	u^{k+1}= \widetilde{G}(\boldsymbol	u^{k})	
\end{equation}
which first recover $({\boldsymbol R_\text X}^{k},{\boldsymbol t_\text X}^{k})$ from $\boldsymbol	u^{k}$, then perform $G$ call in (3), and finally convert $({\boldsymbol R_\text X}^{k+1},{\boldsymbol t_\text X}^{k+1})$ back to $\boldsymbol u^{k+1}$.

In addition, Anderson acceleration is proved to be convergent under mild condition when starting point is close to the solution\cite{toth2015AAslowconvergence}, but the initial guess of hand-eye relation is not always good enough to be adjacent to the fixed point we are looking for. In that case AA may suffer slow convergence or stagnation. With that in mind, we applied heuristics which combines the safe-guarding steps proposed in \cite{pavlov2018AAICP} and \cite{peng2018andersonAAheuristics} to improve its stability. Specifically, we check the mean of squared distance between corresponding points (i.e. MSE) determined by each hand-eye relation iterate and accept the accelerated one as new iterate only if its error is not considerably larger than the previous iterate. Otherwise, we revert to the last trustworthy point given by the last $\widetilde{G}$ call as the new iterate and reset the length of history.

\begin{figure}[!t]
	\centering
	\includegraphics[width=0.5\textwidth]{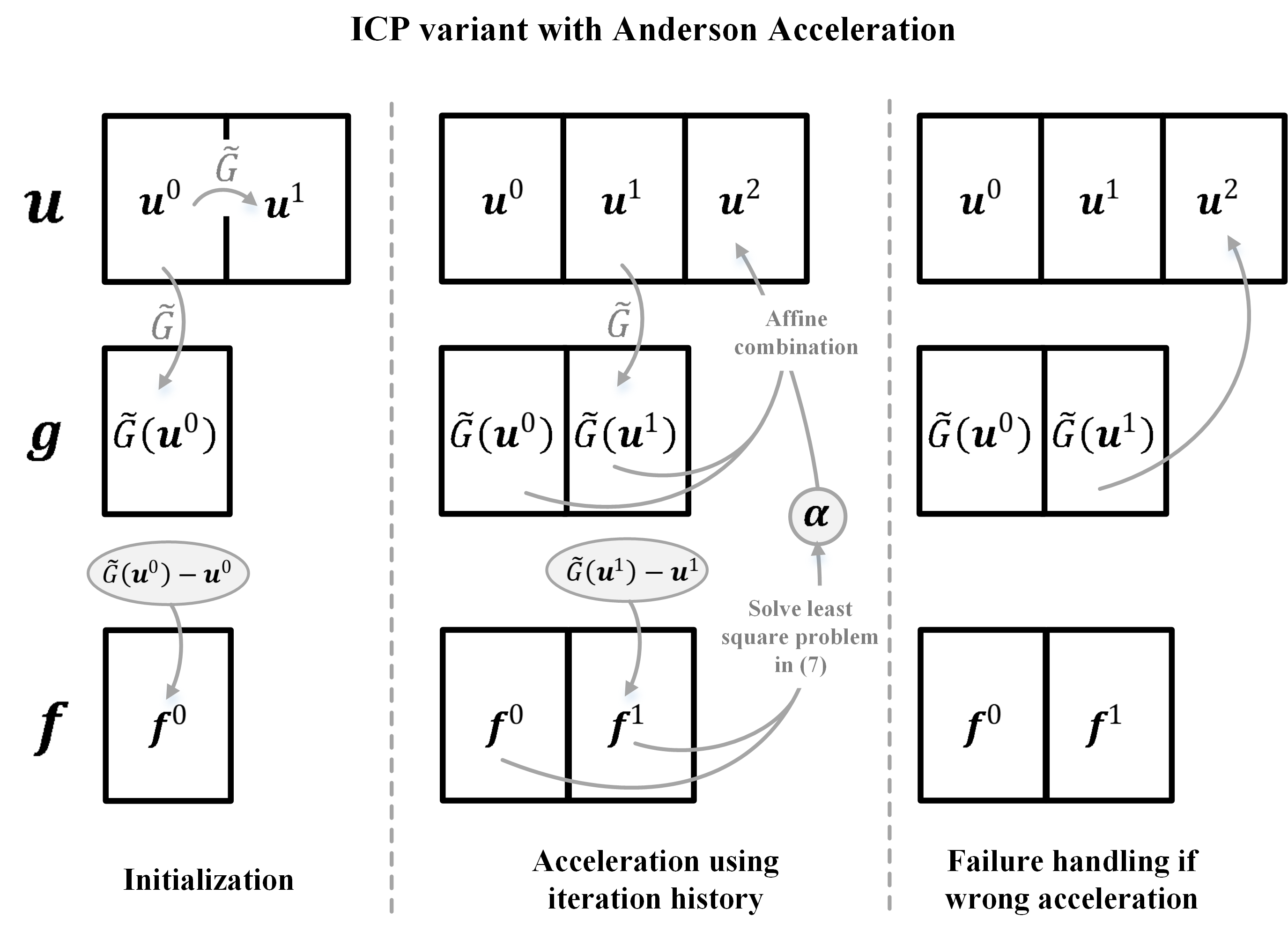}
	\caption{Iteration procedure of ICP variant with Anderson Acceleration.}
	\label{fig_5}
\end{figure}

Fig.5 illustrates the iteration process of ICP variant with Anderson Acceleration. Unlike our previous simple iteration strategy which only applies ${G}$ call repeatedly, the Anderson accelerated ICP variant maintains 3 sequences with $\boldsymbol u$ for hand-eye relation iterates, $\boldsymbol g$ for $\widetilde{G}$ call of the iterates and $\boldsymbol f$ for residual. In the initialization step, we pushback the initial guess of hand-eye relation $\boldsymbol u^0$, its $\widetilde{G}$ call and the residual, then simply push back $\widetilde{G}(\boldsymbol u^0)$ as second accepted iterate $\boldsymbol u^1$. Next for each accepted iterate, pushback its $\widetilde{G}$ call and residual, then calculate the accelerated point according to (5) and (7). This accelerated point will only be accepted if it shows error not considerably larger than the last accepted iterate, otherwise we append the result of last $\widetilde{G}$ call to the iterate sequence $\boldsymbol u$ as the new iterate and reset iteration history. We name this ICP variant with Anderson Acceleration AA-ICPv and pseudocode is given in Algorithm 1. 

\begin{algorithm}
	\small	
	\textbf{Input:} 
	\\ Initial guess of hand-eye relation $\boldsymbol u^0$;\\ Point clouds captured from  $n \!\!+\!\! 1$ different viewpoints and corresponding robot poses, during $n$ robot motions;\\Convergence threshold $\epsilon$; Maximum history length $l$(default=4)\\  
	 \textbf{Define:}\\
	$Correspondence(\boldsymbol u)$: Convert $\boldsymbol u$ into $(\boldsymbol R,\boldsymbol t)$ and transform all point clouds into robot base frame. Then determine all the correspondences $\mathcal{S}$, i.e. $\{({\boldsymbol p_j^i},{\boldsymbol q_j^i})\} $( $ i=1,...,n,j=1,...,m_i$ )\\
	$Alignment(\mathcal{S})$: Find the hand eye relation that minimize the squared distances of $\mathcal{S}$, i.e. solve problem defined in (1), then return the hand-eye relation in $\boldsymbol u$\\
	$E(\boldsymbol u)$: Mean of squared distance (i.e. MSE) of $\mathcal{S}$ determined by $\boldsymbol u$ 
	\caption{AA-ICPv}             	
	\label{algorithm 1: }                        		
	\begin{algorithmic}[1]	
        \STATE $k_{start}=k=0$; $\mathcal{S}^0=Correspondence(\boldsymbol u^0)$;
         $E_{prev} = +\infty$;\\
         \STATE $\boldsymbol g^0 = Align(\mathcal{S}^0)$;$\boldsymbol f^0 = \boldsymbol g^0-\boldsymbol u^0$;$\boldsymbol u^1 = \boldsymbol g^0$;$k = k+1$
    
		\WHILE{\TRUE }
		  \IF{$E(\boldsymbol u^k)>E_{prev}$} 
		  \STATE $k_{start}=k-1$; $\boldsymbol u^k=\boldsymbol g^{k-1}$;$E_{prev} = +\infty$\\
		  \STATE \textbf {Continue};		  
		  \ENDIF
		\STATE $ E_{prev}=E(\boldsymbol u^k)$
		  		  
		  \STATE$\mathcal{S}^k=Correspondence(\boldsymbol u^k)$;$\boldsymbol g^k = Align(\mathcal{S}^k)$
		  \STATE$\boldsymbol f^{k}=\boldsymbol g^{k}-\boldsymbol u^{k}$\\
		  $//\mathrm{Check\quad convergence}$
		  \STATE\textbf{if}  {$||\boldsymbol f^{k}||<\epsilon$}  \textbf{then Break};\\
		  $//\mathrm{Anderson\quad Accleration}$
		  \STATE $m=min(k \!-\! k_{start},l)$
          \STATE $(\alpha_0,...,\alpha_{m-1})=argmin   ||\boldsymbol f^{k}+\sum_{i=0}^{m-1} \alpha_i (\boldsymbol f^{k-m+i}-\boldsymbol f^{k})||_2$
          \STATE $\alpha_m=1-\sum_{i=0}^{m-1}\alpha_i$;$\boldsymbol u^{k+1}=\sum_{i=0}^m\alpha_i \boldsymbol g^{k-m+i}$
          \STATE $k=k+1$

		\ENDWHILE				
	\end{algorithmic}	
\end{algorithm}

\begin{figure}[!t]
	\centering
	\includegraphics[width=0.5\textwidth]{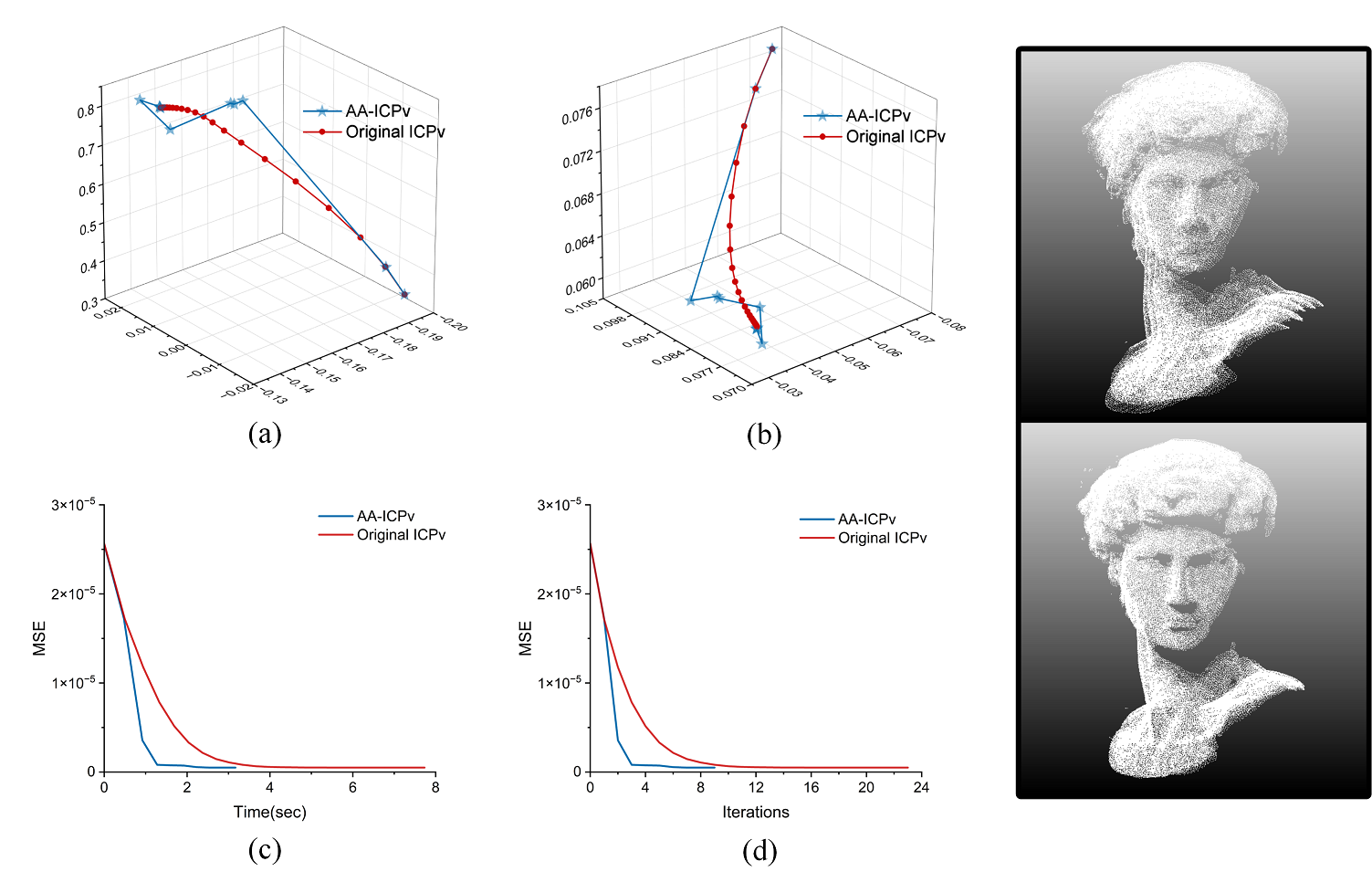}
	\caption{A comparison between iterate sequence of original ICPv and AA-ICPv, using same initial hand-eye relation. (a)Rotation sequence in so(3). (b)Translation sequence in 3-D Euclidean space. (c)MSE over time. (d)MSE over iterations. Point cloud above and below demonstrate the registration performance of original ICPv and AA-ICPv respectively, both at 4th iteration.}
	\label{fig_6}
\end{figure}
\begin{figure}[!t]
	\centering
	\includegraphics[width=0.5\textwidth]{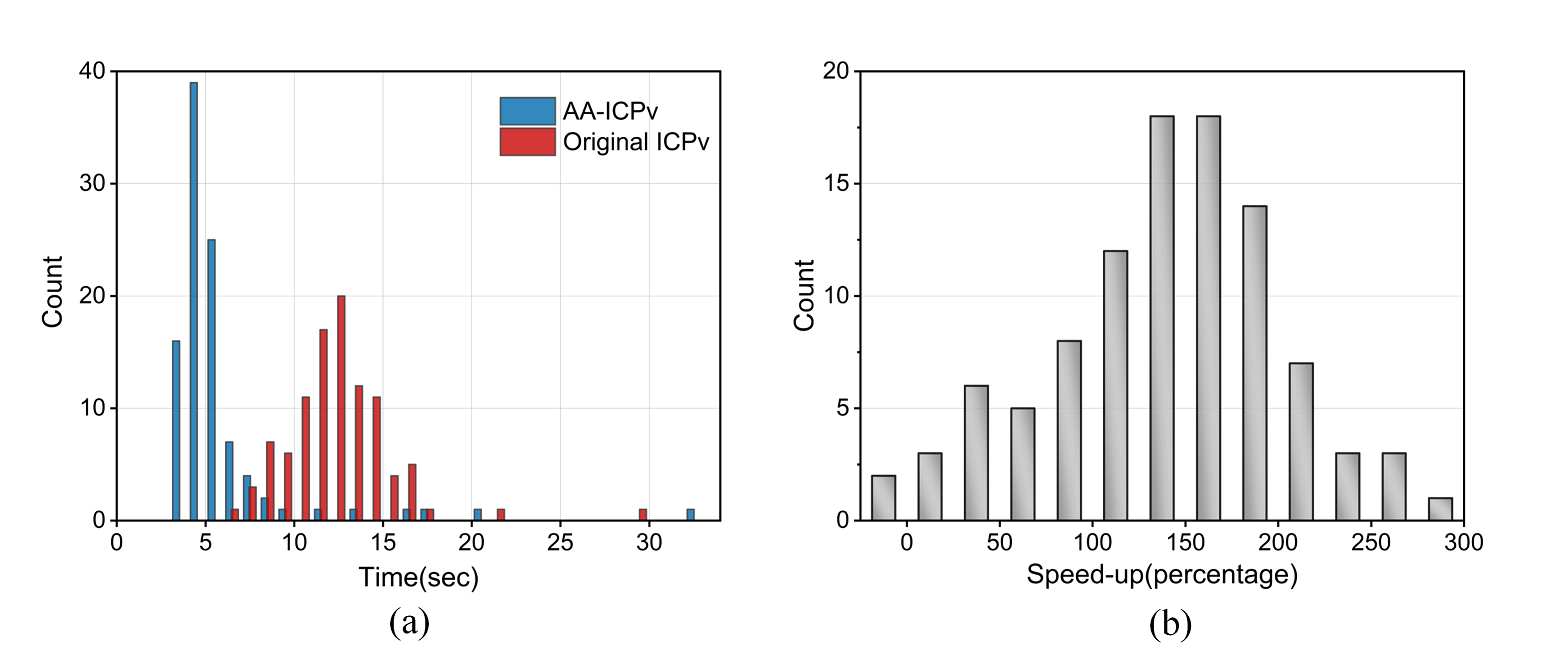}
	\caption{Statistical comparison between original ICP and AA-ICPv on convergence speed. (a)Histogram of time required for convergene. (b)Histogram of speed-up in percentage.}
	\label{fig_7}
\end{figure}

To test the effectiveness of Anderson Acceleration, 9 point clouds of a David plaster figure were taken by eye-in-hand system from different viewpoints. An comparison between original ICPv and AA-ICPv on iteration procedure with same initial guess is given in Fig.6.
For easy visualization, we separate the iterate sequence into a pure rotation sequence (first three elements of $\boldsymbol u$) and a pure translation sequence (last three elements of $\boldsymbol u$), then plot the two sequences into so(3) space (shown in subplot (a)) and 3-D Euclidean space (shown in subplot (b)) respectively. Corresponding MSE behavior over time and iterations are given in subplot (c) and (d). One can see that, both iterate sequence start from the same initial guess and converge to same fixed point. Unlike the Cauchy sequence generated by original ICPv,
the jumpy nature of Anderson Acceleration enables faster convergence with less iterations. A comparison in registration performance, both on 4th iteration, is shown on the right of Fig.6.

Above experiment was repeated with 100 different initial guess and the histogram of time required for convergence is plotted in Fig.7(a). Fig.7(b) illustrate the speed-up ratio in histogram. Overall, 98\% runs got accelerated relative to original ICP variant and 76\% runs got accelerated more than 100\%. Median speed-up percentage and mean speed-up percentage are \%147 and \%138 respectively.

\subsection{Bayesian optimization based Initial alignment (BO-IA)}
Like classic pair wise ICP, the proposed ICP variant with Anderson Acceleration can be trapped into local minimum given poor initial guess, leading to erroneous registration and vastly different calibration result. A rational initial-guess of hand-eye relation is then required to avoid local minimum during convergence and wrong acceleration during AA. Although it is promising to manually input an initial guess according to the design CAD model of sensor and senor-holding structure\cite{murali2021ManualInput}, there is still great appeal for automatic approaches that can provide rough estimation of hand-eye relation and bring multi-view point clouds into coarse registration.

We noticed that the error function $E(\boldsymbol u)$ given in last subsection, which returns the MSE of estimated correspondences determined by $\boldsymbol u$, can be taken as a metric of registration performance. In the meanwhile, it is hard to differentiate and non-convex, making it an ideal candidate for Bayesian optimization(BO), a probability-driven derivative-free global optimization technique reported to be instrumental in many robotics-related research\cite{cully2015robotsBOrobotics1,chatzilygeroudis2018BOrobitics2}. Thus we modeled the registration error over hand-eye relation in a continuous search space as Gaussian Process(GP), and carry out the following processes alternately:(1)update the posterior according to already sampled point and prior. (2)decide where to sample next according to an acquisition function and updated posterior.

In RegHEC, we keep the parameterization $\boldsymbol u$ for hand-eye transformation, as it's an compact representation with no redundancy and constraints, favorable for Bayesian optimization\cite{frazier2018tutorialBO}. We define the search space as a 6-dimensional hyper-rectangle, where the rotation domain is the minimum cube $[-\pi,\pi]^3$ that encloses the $\pi$-ball and the translation domain is a roughly estimated cube that encloses the sensor.

The prior distribution of GP before any observations is defined by a mean function $\mu_0(\boldsymbol u)$ and covariance function $k(\boldsymbol u_1,\boldsymbol u_2)$(also known as kernel function). Assuming all points in the search space are equally likely to be good, the mean vector is initialized with a constant. Based on the assumption that objective function is continuous and smooth, the kernel is designed so that nearby points $\boldsymbol u_1, \boldsymbol u_2$ in the search space have a large positive correlation, encoding the belief that they should have more similar error function value than points that are far apart. 

Given observations $E(\boldsymbol u_{1:n})$ at $n$ random sample points $\boldsymbol u_{1:n}$, where $E(\boldsymbol u_{1:n})\!=\!{[E(\boldsymbol u_1),...,E(\boldsymbol u_n)]}^{\text T}$ and $\boldsymbol u_{1:n}$ indicates the sequence $\boldsymbol u_1$,...,$\boldsymbol u_n$,  then  posterior distribution of error function value at $\boldsymbol u$ is updated as
\begin{equation}
	E(\boldsymbol u)|E(\boldsymbol u_{1:n}) \sim \mathcal{N}(\mu_n(\boldsymbol u),\,\sigma_n^{2}(\boldsymbol u))
\end{equation}
with
\begin{equation}
	\mu_n(\boldsymbol u)=\boldsymbol k\boldsymbol K^{-1}(E(\boldsymbol u_{1:n})-\mu_0(\boldsymbol u_{1:n}))+\mu_0(\boldsymbol u)
\end{equation}
\begin{equation}
	\sigma_n^{2}(\boldsymbol u)=k(\boldsymbol u,\boldsymbol u)-\boldsymbol k \boldsymbol K^{-1} \boldsymbol k^{\text T}
\end{equation}
where
\begin{equation}
	\boldsymbol K=\begin{bmatrix}
		k(\boldsymbol u_1,\boldsymbol u_1)&\cdots&k(\boldsymbol u_1,\boldsymbol u_n)\\ 
		\vdots&\ddots&\vdots\\
		k(\boldsymbol u_n,\boldsymbol u_1)&\cdots&k(\boldsymbol u_n,\boldsymbol u_n)		
	\end{bmatrix}
\end{equation}
\begin{equation}
	\boldsymbol k=\begin{bmatrix}
		k(\boldsymbol u,\boldsymbol u_1)&\cdots&k(\boldsymbol u,\boldsymbol u_n) 
	\end{bmatrix}
\end{equation}

According to the posterior distribution, where to sample next is decided by maximization of an acquisition function. We apply expected improvement metric(EI), which tells where the optimal point is likely to be based on tradeoff between exploration and exploitation.
\begin{equation}
	EI_n(\boldsymbol u)=\int_{-\infty}^{+\infty}max(E_n^*-E(\boldsymbol u),0)p(E(\boldsymbol u)|E(\boldsymbol u_{1:n}))dE(\boldsymbol u)
\end{equation}
where $E_n^*$ is the best observation in the already sampled $n$ points thus far. Next point to be sampled is then decided as 
\begin{equation}
	\boldsymbol u_{n+1}=argmaxEI_n(\boldsymbol u)
\end{equation}

The prior setting is critical for the Gaussian process, which affects the efficiency of Bayesian optimization. We take the mean of already sampled function value as the prior mean of GP. However, the commonly used covariance function like squared exponential kernel(Exp), Matern kernel and squared exponential with auto correlation detection(ExpARD) are not applicable for optimization over 3-D motion space SE(3) parameterized by $\boldsymbol u$, due to the incompatible distance metric. For instance, $\boldsymbol u_1(0,0,-\pi,0,0,0)$ and $\boldsymbol u_2(0,0,\pi,0,0,0)$ can be far apart under Euclidean distance which is applied by square exponential kernel, but they are essentially same hand-eye relation with no difference in corresponding MSE. Given sample point at $\boldsymbol u_1$, squared exponential kernel will generate a large posterior uncertainty for function value at $\boldsymbol u_2$, and acquisition function may decide to sample nearby next. To avoid above futile sampling, we proposed a new covariance function encoding an effective distance metric, trying to provide correct correlation for function value over 3-D motion space SE(3) parameterized by $\boldsymbol u$.   
\begin{equation}
	k_{SE(3)}(\boldsymbol u_1,\boldsymbol u_2)=\sigma ^2exp(-\frac{d(\boldsymbol u_1,\boldsymbol u_2)}{2l^2})
\end{equation}
with
\begin{equation}
	d(\boldsymbol u_1,\boldsymbol u_2)=||log({exp(\boldsymbol v_1^\wedge)}^{-1}exp(\boldsymbol v_2^\wedge))||+\alpha^2||\boldsymbol t_1-\boldsymbol t_2||
\end{equation}
where ${(\cdot)}^\wedge$ means skew symmetric matrix and $exp(\cdot^\wedge)$ is exponential mapping from so(3) to SO(3)\cite{sola2018microLie}.

It is similar to squared exponential kernel but deploys a more meaningful and effective distance metric, which is the sum of rotation distance for rotation part and Euclidean distance multiplied by some scale factor for translation part. The above hyperparameters $\sigma$,$l$ and $\alpha$ are optimized with Maximum Likelihood Estimation(MLE)\cite{Blum2013OBOhpopt}. That is, for the given sampled observations, we calculate the likelihood of those observations under prior determined by hyperparameters, then set hyperparameters to the value that maximizes this likelihood.

\begin{figure}[!t]
	\centering
	\includegraphics[width=0.48\textwidth]{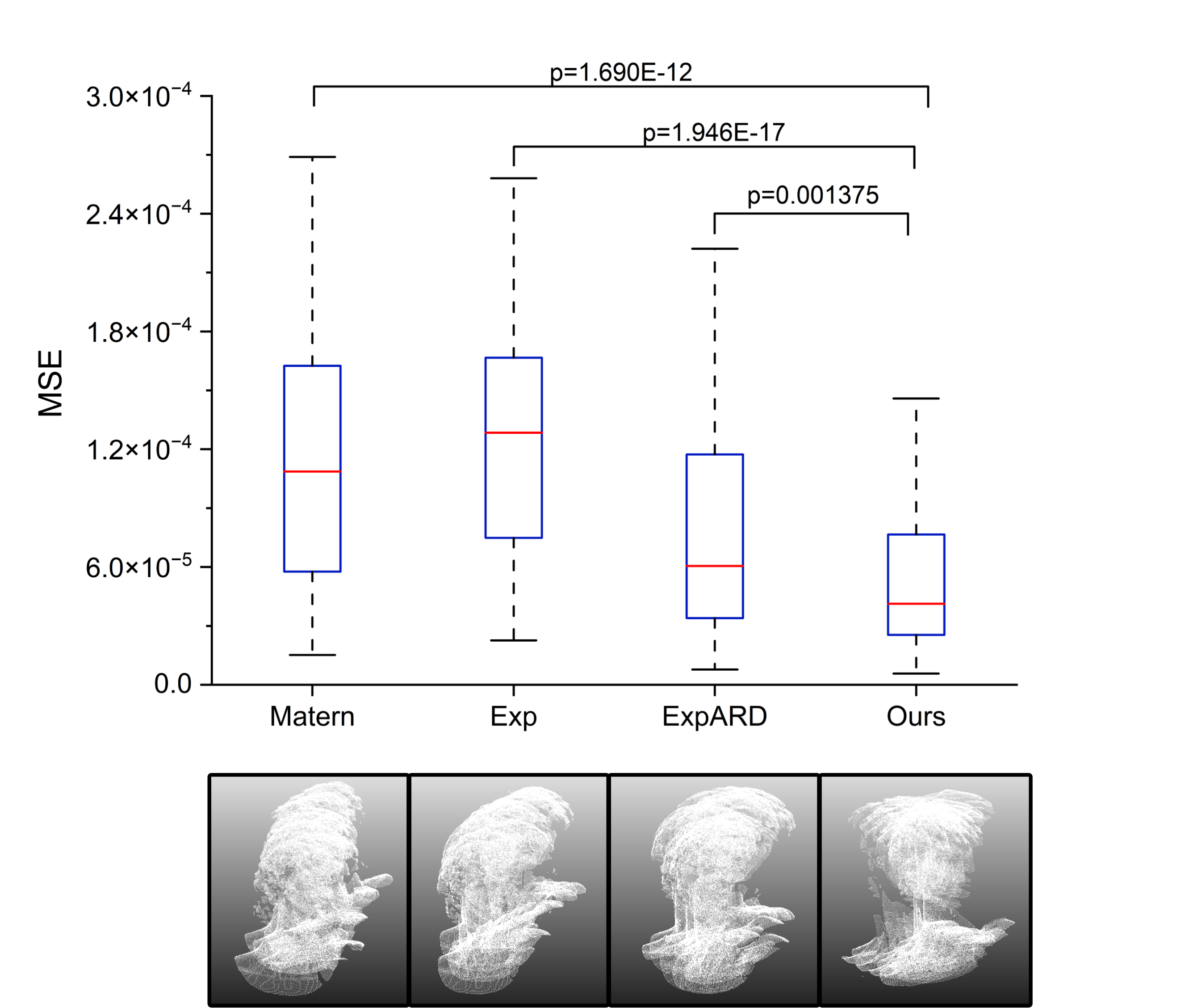}
	\caption{A statistical comparison between  different kernel on initial alignment performance. Each box denotes 100 rounds of test. Point clouds below help to visualize the initial alignment with median result of each kernel.}
	\label{fig_8}
\end{figure}

RegHEC first make some initial random observations, then iteratively alternate between posterior update and new point sampling. It returns the point with best observation as initial guess of hand-eye relation when total number of sampling is reached. As the process of solving the initial transformation is usually called initial alignment or coarse registration, we name this method Bayesian optimization based initial alignment(BO-IA). Please refer to Algorithm 2 for more detail.
\begin{algorithm}
	\small
	\textbf{Input:} Point clouds captured from different viewpoints and corresponding robot poses;\\Search space $\mathrm{b}$; Number of initial random sampling $n_0$(default=50); Number of total sampling $N$(default=100); Period of hyper paratmerer optimization $p$(default=10)\\
	\textbf{Output:} Initial guess of hand-eye relation and coarse registration of multi-view point clouds.
	\caption{BO-IA}             	
	\label{algorithm 3:}                        		
	\begin{algorithmic}[1]	
		\STATE $n=n_0$		
		\STATE Observe $E(\boldsymbol u)$ at $n$ random points $\boldsymbol u_{1:n}$ within $\mathrm{b}$.						
		\WHILE{$n<N$ }
		\STATE $n=n+1$
		\IF{$(n-n_0)\%p==0$}
		\STATE Optimize kernel hyper parameters with MLE
		\ENDIF		
        \STATE Update posterior distribution on $E(\boldsymbol u)$ with already sampled $n$ points
		\STATE Find the maximizer of EI as next point to be sampled $\boldsymbol u_{n+1}$ within $\mathrm{b}$ then observe $E(\boldsymbol u_{n+1})$				
		\ENDWHILE      
		\RETURN Point evaluated with best $E(\boldsymbol u)$
	\end{algorithmic}	
\end{algorithm}

With the 9 point clouds of David plaster figure used in previous subsection, performance of proposed covariance function was compared to Exp, ExpARD and Matern kernel. We ran the BO-IA algorithm 100 times with each of the covariance function while keeping all other parameters the same. MSE of best observation of all these runs are given in Fig.8. Point clouds initial alignment corresponding to result at median of each 100 runs are presented below the box plot, for easy visualization. Two-tail t-tests were conducted to compare the performance of proposed kernel versus that of the others. The proposed kernel achieves significantly better initial alignment than others at level of 5\%, implying that it is able to provide a more rational GP for BO over continuous 3-D motion space SE(3) parameterized by $\boldsymbol u$.

\subsection{Extension to Eye-to-hand scenario}
\begin{figure}[!t]
	\centering
	\includegraphics[width=0.5\textwidth]{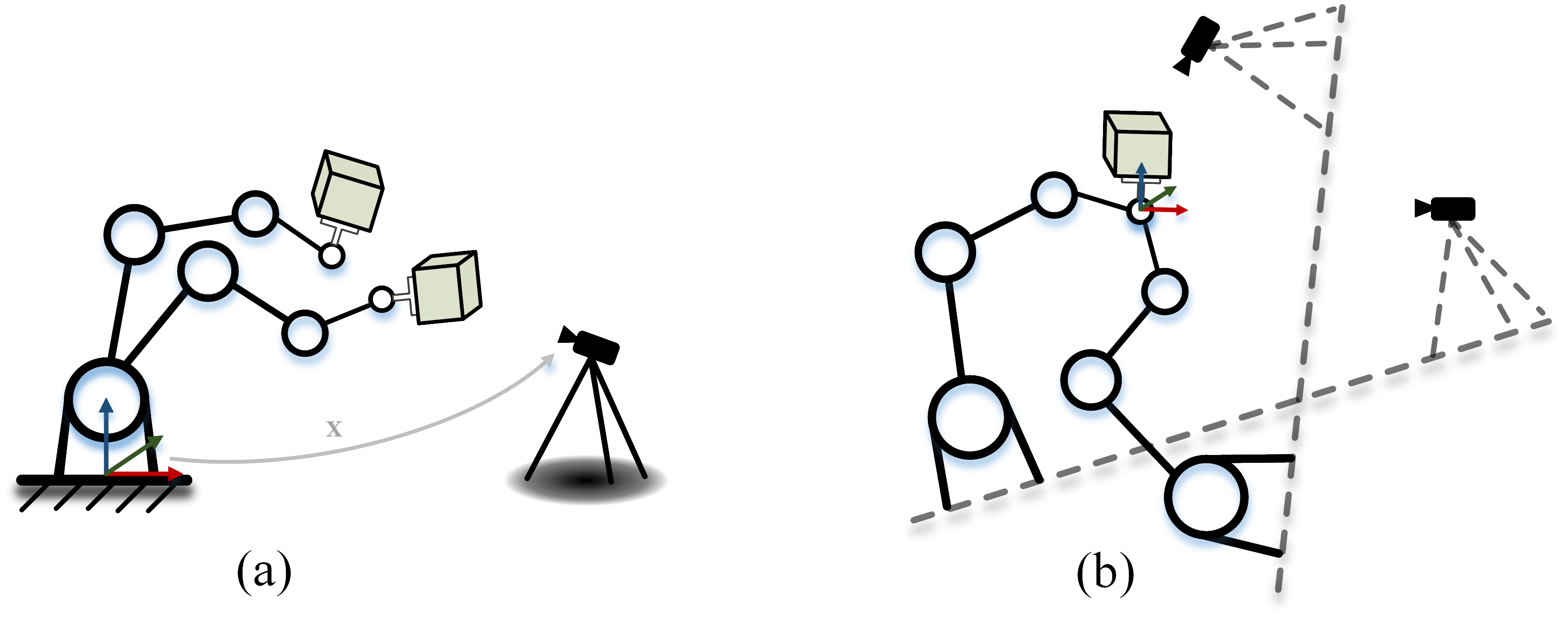}
	\caption{By switching reference frame from robot base(a) to robot flange(b), RegHEC can be easily extended to eye-to-hand scenario. Consider the robot flange as "base" and base as "flange", then plug in the inverse of robot poses, multi-view point clouds registration of object in hand and base-sensor calibration can be achieved simultaneously.}
	\label{fig_9}
\end{figure} 
For eye-to-hand scenario where transformation between sensor and robot base is to be calibrated, people tend to set the reference frame at robot base or sensor, then consider the sensor as stationary and end effector/object in hand as moving, as is shown in Fig.9(a). 

However, change of the reference frame won’t affect the relative pose between end effector/object in hand and sensor, thus one can also see a flying sensor and stationary end effector/object in hand, by setting the reference frame at the robot flange. It follows that robot flange and base are essentially the robot "base" and "flange" in eye-in-hand case respectively, as is shown in Fig.9(b). Apparently, multi-view point clouds of end effector/object-in-hand align after transformation into robot flange frame using inverse of corresponding robot poses and base-sensor relation. This suggests that we can control the robot to present different angles of the object in the hand to the stationary sensor for point clouds acquisition, then simultaneous base-sensor calibration and multi-view point clouds registration can be achieved via RegHEC.

In addition, if one uses end effector only(empty hand) as reconstruction target, the TCP of end effector can be calibrated in the meantime, as registration is under robot flange frame.

\subsection{Some implementation details}
We name the resulting algorithm RegHEC, where BO-IA achieves coarse registration then provide the proper initial guess of hand-eye relation to AA-ICPv for later fine registration and accurate calibration. Some dependencies in the released C++ implementation are PCL\cite{PCL} for point clouds operation, Limbo\cite{cully2018limbo} for BO and Sophus for Lie algebra calculation.
 
As mentioned earlier, the robot is required to pauses at $n+1$ poses during $n$ motions, and a target point cloud is captured at each robot poses. Thus we have in total $n$ pairs of point clouds by pairing point clouds captured before and after each motion. During correspondences step, we go through all $n$ pairs of point clouds. In each pair, for each point in the smaller point cloud, we find its corresponding point(closest) in the other point cloud to form one correspondence, then discard those correspondences with distance over certain threshold. A commonly used distance threshold is constant, whereas too large a threshold will include too many outliers thus leading to erroneous final registration and too small a threshold will make convergence difficult as many inliers are excluded. The initial guess given by BO-IA is random (see Fig.8), thus setting a universal threshold, applicable for all possible cases, is almost impossible. Inspired by the Trimmed ICP\cite{chetverikov2005TRICP1,chetverikov2002trimmedICPfailsmmetrical} for robust registration, we implemented a dynamic distance threshold defined by a trimming ratio $\eta$. Specifically, we sort the distances between 2 points of all the correspondences in an increasing order, and only keep the $k$ correspondences with least value, where $k$ equals ratio $\eta$ times total number of correspondences.

When solving the least square problem using algorithm 3(see Appendix) after correspondences are decided, we update the hand-eye relation only once, that is, we do not iterate until update step $\Delta X$ is small enough but only calculate $\Delta X$ once, for following reasons. First, during the early iterations in the accelerated ICP variant, the matched correspondences are usually incorrect. The first step in Gauss-Newton usually brings more decrease in the objective than later update step, it significantly drags point clouds closer to each other just as expected. Second, during the late iterations in the accelerated ICP variant(close to convergence), the optimum to be solved in $\widetilde{G}$ call is very close to current iterate, thus one step is usually sufficient to provide solution good enough. Solving the normal equation is time-consuming and this implementation makes RegHEC faster with little loss in accuracy.

Furthermore, the BO-IA and trimming distance calculation use downsampled subset of input point clouds for faster computation.
 
\section{Experiments}

\subsection{System setup and calibration}
\begin{figure}[!t]
	\centering
	\includegraphics[width=0.5\textwidth]{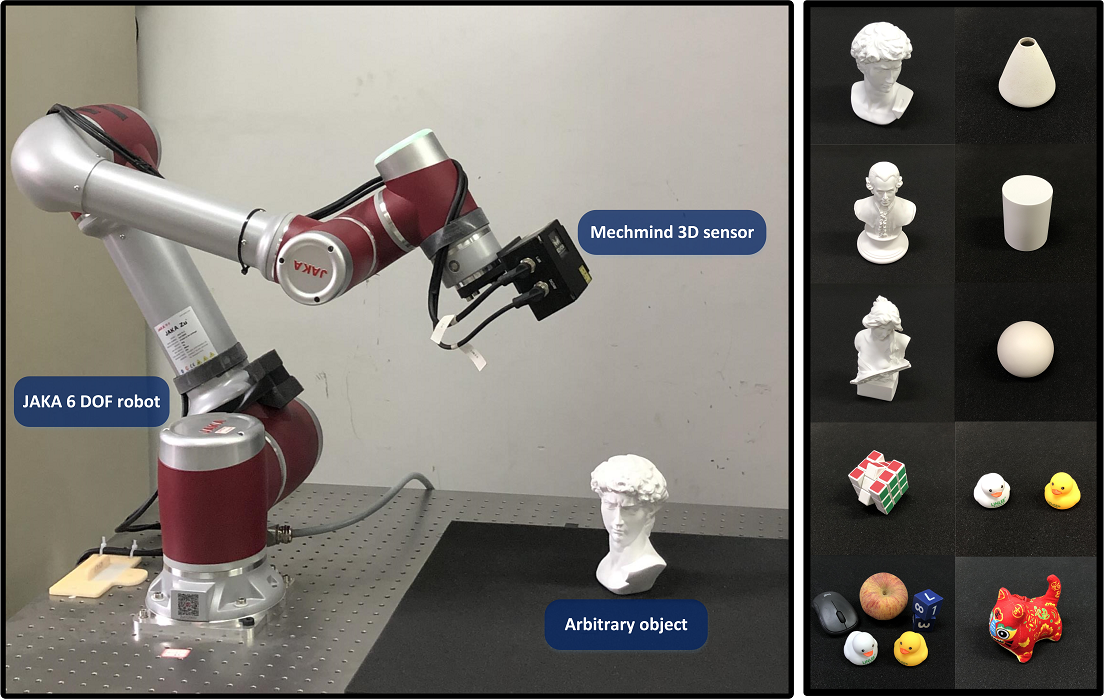}
	\caption{Eye-in-hand experiment setup. A 6-DOF serial manipulator carrying a snapshot 3-D sensor captures point clouds of a stationery scene from different viewpoints. Objects used for experiment are shown on the right. From top to bottom, left column shows David(plaster figure), Mozart, nymph, Rubik and object cluster. Right column shows cone, cylinder, sphere, rubber ducks and tiger mascot.}
	\label{fig_10}
\end{figure}
\begin{figure*}[!t]
	\centering
	\includegraphics[width=0.85\textwidth]{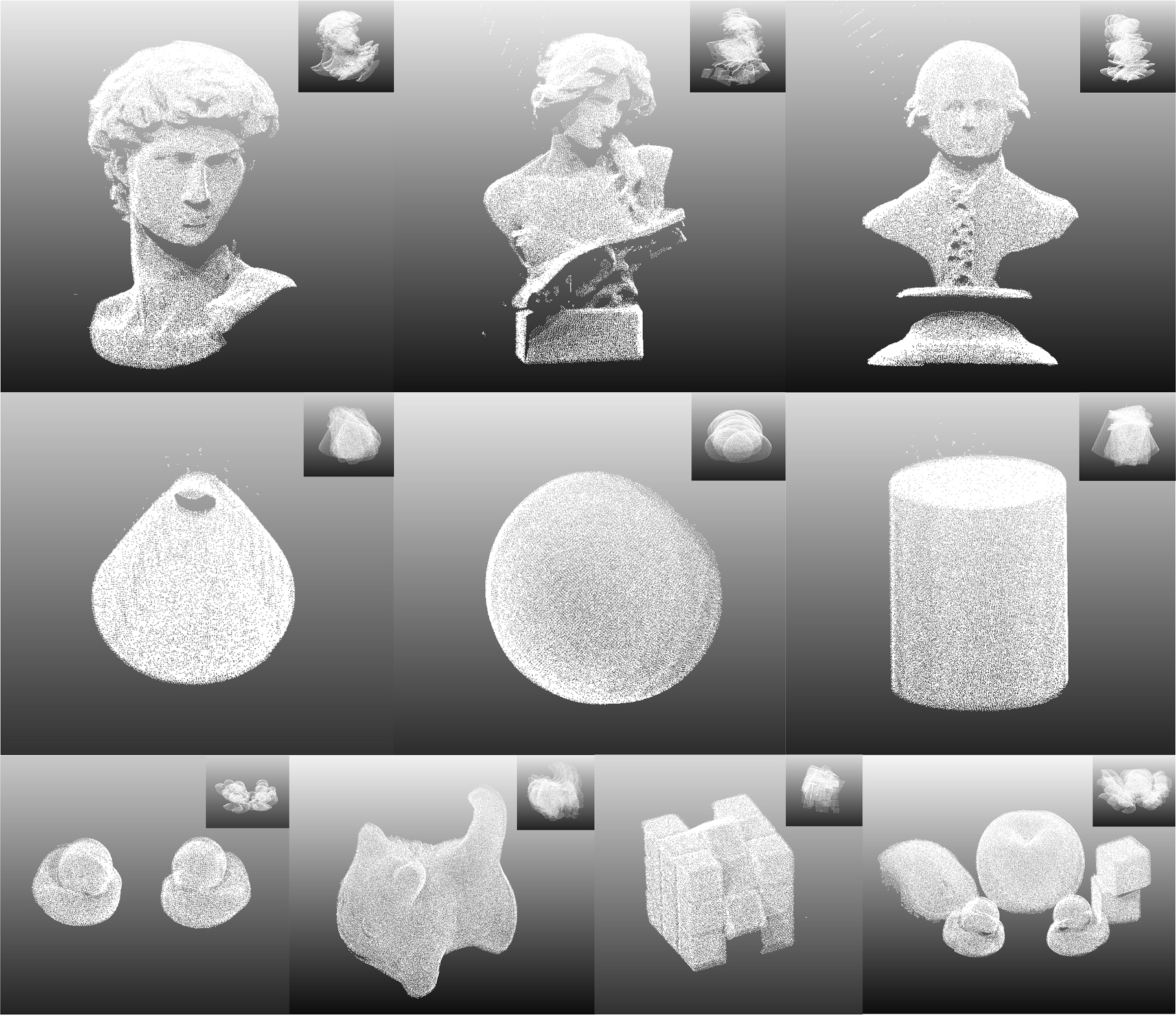}
	\caption{Registration of 9 point clouds from different viewpoints under robot base frame, with RegHEC, when convergence. Miniature in the top right shows the 9 source point clouds under sensor frame before registration.}
	\label{fig_11}
\end{figure*}
A hand eye system consisting of a 6-DOF serial manipulator and a snapshot 3-D sensor is established. We first mounted the sensor on the flange of the manipulator to test the performance of RegHEC in eye-in-hand scenario, as is shown in Fig.10. Varieties of arbitrary  objects with no prior information were used during the experiment, including plaster figures(David, nymph and Mozart), Rubik, rubber ducks, tiger mascot and an object cluster. Besides, three SORs including cylinder, cone and sphere were also utilized, which can be challenging for point cloud registration and sensor motion estimation with conventional pair-wise ICP. 

For each of the 10 scenes, point clouds were captured from 9 different viewpoints and downsampled using voxel grid filter with leaf size of 0.001m. Corresponding robot poses were read from robot controller then recorded.

For each of possible combination of scenes and number of point clouds used(from 3 to maximum 9), we run RegHEC 100 times on a standard laptop featuring an Intel Core i7 9750H CPU and 16 GB of memory, with parameters set as follows: For BO-IA, we set the search space $\mathrm{b}$ as $[-\pi,\pi]^3\times[-0.1,0.1]^3$. For AA-ICPv, the convergence threshold $\epsilon$ was set as 0.0001 and trimming ratio $\eta$ for dynamic distance threshold computation was set to be 0.9. Among the total 7000 runs, 98.714\% converged within 100 iterations and Fig.11 shows examples of 9 point clouds of different scenes before and after registration using RegHEC.

\subsection{Accuracy assessment}
\begin{figure}[!t]
	\centering
	\includegraphics[width=0.5\textwidth]{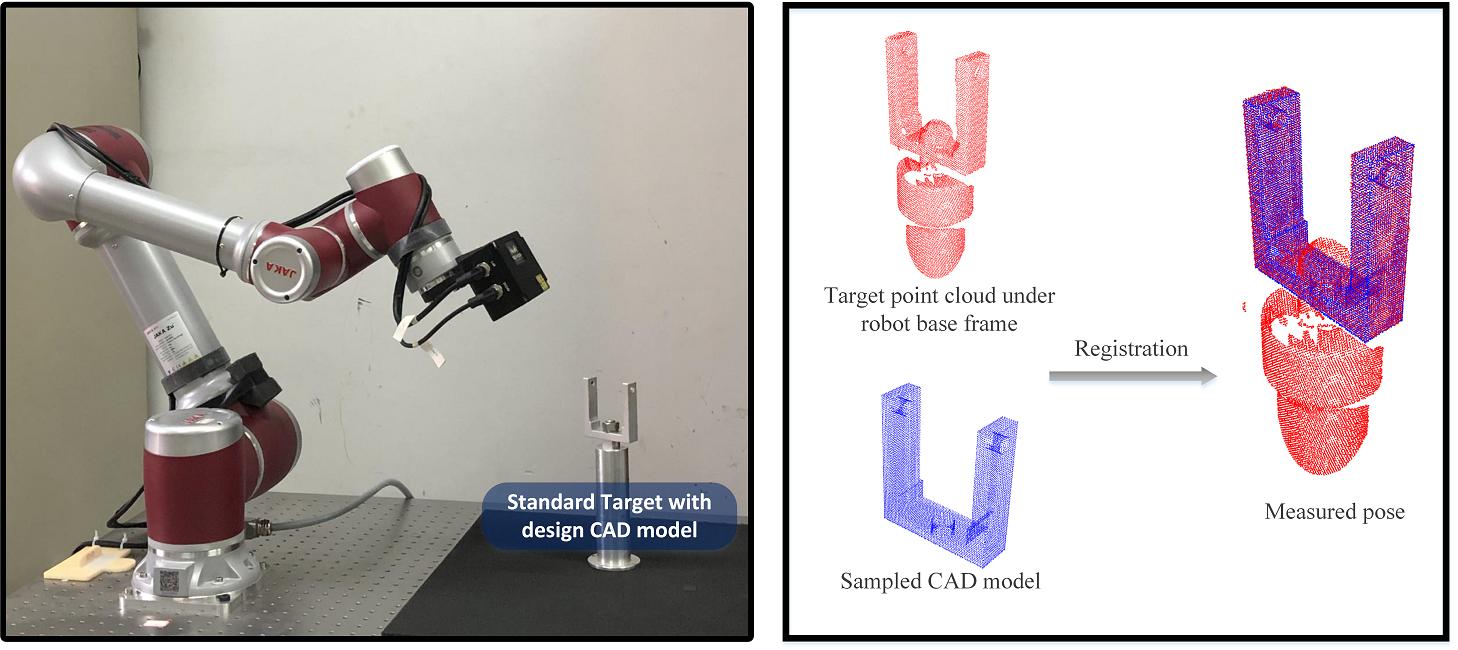}
	\caption{The pose measurement experiment for accuracy assessment. Left: Test data acquisition. Right: Point cloud registration for pose measurement.}
	\label{fig_12}
\end{figure} 
It is widely acknowledged that the ground truth of hand-eye relation is hardly available. Considering that point clouds need to be transformed to robot base for alignment during robotic 3-D reconstruction task, a pose measurement experiment was designed for accuracy evaluation from application perspective. Specifically, we captured point clouds of a stationary standard mechanical target with known CAD model(see Fig.12) from 50 random viewpoints different than those utilized for calibration. Next we transformed the 50 point clouds into robot base frame using corresponding robot poses and hand-eye relation to be assessed. Then pose of each 3-D scan, w.r.t. robot base frame, was measured with the sampled CAD model via pair-wise registration.

\begin{table*}[h!t]	
	\renewcommand\arraystretch{1.32}
	\small
	\centering
	\caption{{Accuracy assessment ${Err}_{\boldsymbol R}$/${Err}_{\boldsymbol t}$ in eye-in-hand scenario. (unit:{$^\circ$/\upshape mm})}}
	\label{tab4}
	\resizebox{1.95\columnwidth}{!}{      
		\begin{tabular}{c c c c c c c c c }\hline
			\multirow{2}*{Method}& \multirow{2}*{Objects}& \multicolumn{7}{c}{Numder of input point clouds/images} \\
			\cline{3-9} 
			&\qquad&3&4&5 &6 &7 &8 &9 \\ \hline
			RegHEC&David\qquad&0.733/2.496 &0.396/0.702 &0.335/0.653 &0.361/0.795 &0.263/0.593 &0.273/0.602 &0.299/0.627 \\
			RegHEC&Nymph\qquad&1.341/3.605 &0.297/0.735 &0.294/0.740 &0.234/0.832 &0.136/0.458 &0.149/0.501 &0.167/0.515 \\	
			RegHEC&Mozart\qquad&1.421/4.415 &0.360/0.650 &0.305/0.614&0.319/0.750 &0.172/0.440 &0.209/0.492&0.226/0.502\\
			RegHEC&Rubik\qquad&1.392/3.445 &0.224/0.736 &0.219/0.684 &0.162/0.426 &0.258/0.619 &0.189/0.524 &0.175/0.478\\		
			RegHEC&Cluster\qquad&0.436/1.137 &0.194/0.683 &0.268/0.620 &0.205/0.583 &0.166/0.504 &0.159/0.502 &0.179/0.536\\
			RegHEC&Cone\qquad&3.563/5.179&0.447/1.018&0.361/0.846&0.260/0.669&0.534/0.873&0.174/0.510&0.169/0.503\\
			RegHEC&Sphere\qquad&6.868/12.024 &1.099/1.838 &0.383/0.815 &0.457/1.015 &0.250/0.676 &0.266/0.697 &0.320/0.744\\
			RegHEC&Cylinder\qquad&4.645/7.307 &0.234/0.782 &0.370/0.781 &0.461/1.092 &0.138/0.540 &0.167/0.896 &0.227/0.680\\
			RegHEC&Mascot\qquad&1.067/2.431&0.160/0.638&0.173/0.521&0.160/0.482&0.244/0.468&0.197/0.414&0.182/0.384\\
			RegHEC&Duck\qquad&0.490/1.103 &0.210/0.999 &0.266/0.716 &0.229/0.668 &0.180/0.467 &0.149/0.463 &0.160/0.488\\
			Tsai\cite{Tsai}&Calibration board\qquad&0.445/0.772&0.293/0.826 &0.288/0.776 &0.213/0.649 &0.201/0.646 &0.194/0.589 &0.188/0.535\\
			Horaud\cite{HoraudSimulIterative}&Calibration board\qquad&0.453/0.800 &0.310/1.059 &0.299/1.116 &0.256/0.993 &0.254/0.755 &0.238/0.795 &0.230/0.766\\
			Shah\cite{shah2013AXYBseparate}&Calibration board\qquad&0.458/1.938 &0.298/0.779 &0.292/0.758 &0.221/0.667 &0.208/0.598 &0.200/0.565 &0.194/0.509\\
			Li\cite{li2010AXYBsimul}&Calibration board\qquad&0.468/1.324 &0.298/0.967 &0.292/0.836 &0.220/2.434 &0.208/2.663 &0.201/2.864 &0.194/1.821\\
			Tabb\cite{tabb2017AXYBreproj}&Calibration board\qquad&3.438/5.724 &0.463/1.550 &0.258/1.481 &0.405/1.369 &0.283/1.020 &0.222/1.080 &0.220/0.985\\
			J.Wu\cite{wu2019handeyeSO4}&Calibration board\qquad&0.476/1.234 &0.310/0.901 &0.300/0.944 &0.256/0.749 &0.254/0.706 &0.240/0.729 &0.232/0.689\\
			L.Wu\cite{wu2016handeyeTmech}&Calibration board\qquad&0.430/4.154 &0.294/2.786 &0.288/2.558 &0.218/1.060 &0.206/1.118 &0.198/0.984 &0.192/0.954 \\
			
			\hline
	\end{tabular}  }                 
\end{table*}

Ideally, with correct hand-eye relation, the 50 measurement results of the stationary target are consistent, thus we take the standard deviation (SD) of the measured rotations and positions as the error metric, defined as
\begin{equation}
	\begin{small}
		{Err}_{\boldsymbol R}=\sqrt{\dfrac{1}{n}\sum_{i=1}^n{||log(\boldsymbol R_i^\mathrm{T}\tilde{\boldsymbol R})||}^2}
	\end{small}
\end{equation}
\begin{equation}
	\begin{small}
		\!\!\!\!\!\!\!\!\!\!\!\!\!\!	{Err}_{\boldsymbol t}=\sqrt{\dfrac{1}{n}\sum_{i=1}^n{||\boldsymbol t_i-\tilde{\boldsymbol t}||}^2}
	\end{small}
\end{equation}
where $n$ equals 50. The measurement result of rotation and position of $i$th scan are denoted as $\boldsymbol R_i$ and $\boldsymbol t_i$ respectively. $\tilde{\boldsymbol R}$ and $\tilde{\boldsymbol t}$ specify the rotation and position averaging\cite{hartley} of all the 50 measured poses.

From robotic 3-D reconstruction perspective, hand-eye relation with lower ${Err}_{\boldsymbol R}$ and ${Err}_{\boldsymbol t}$ means point clouds transformed to robot base can be better aligned, it helps to facilitate the post processing for better reconstruction, for instance, registration fine tuning with classic ICP to mitigate impact of robot positioning and calibration error\cite{kriegel2015efficientAdditionalICPneeded,wu2014eye2hand1}.

All the 7000 calibrated hand-eye relations went through the assessment and results(median) were tabulated in Table I.

In order to better demonstrate accuracy performance of RegHEC, we additionally carried out a series of hand-eye calibration using other commonly used methods and a $12 \times 9$ calibration checkerboard with side length of 10 mm($\pm$0.01 mm). The same 9 robot poses were used for data collection. We also tabulate these calibration results into Table I to provide a clear comparison.

One can see that, in general, for all ten scenes with arbitrary objects, the calibration accuracy exhibits a increasing trend as number of input point clouds increases. Although RegHEC's accuracy is lower than other methods when only 3 input point clouds, we see almost no difference and even higher accuracy when more and more point clouds were used for calibration. In addition, the three SORs, especially the sphere, present much worse results when small number of input point clouds. We assume that BO-IA finds it difficult to provide proper initial guess and AA-ICPv can be easily trapped into local minimum. Since these objects
possess identical local feature and point clouds are able to revolve about the axis of revolution while keeping aligned. Moreover, correspondences in 3 point clouds only can be insufficient constraints for initial hand-eye relation estimation given these point clouds without distinctive 3-D feature, which makes things even worse and leads to erroneous calibration result. This phenomenon is significantly mitigated by taking more point clouds into calibration, where RegHEC with the three SORs shows similar accuracy compared with other objects/methods.

While quite challenging for existing pair-wise registration methods\cite{chetverikov2002trimmedICPfailsmmetrical,li2021automaticICPfailsymmetrical}, SOR is no problem for RegHEC owing to the constraints introduced by robot absolute positioning. Although
the correspondences estimation is generally incorrect at initial stage of AA-ICPv, it brings the multi-view point clouds close to each other thus refines the hand–eye relation, which, in return, place the multi-view point clouds in more correct positions and more correct correspondences are established in the next iteration. 

Usually, regardless calibration or reconstruction, data from multiple viewpoints is mandatory, thus although RegHEC presents unsatisfying results when using small number of point clouds, it poses little challenge to real application.

\subsection{Runtime}

\begin{figure*}[!t]
	\centering
	\includegraphics[width=0.88\textwidth]{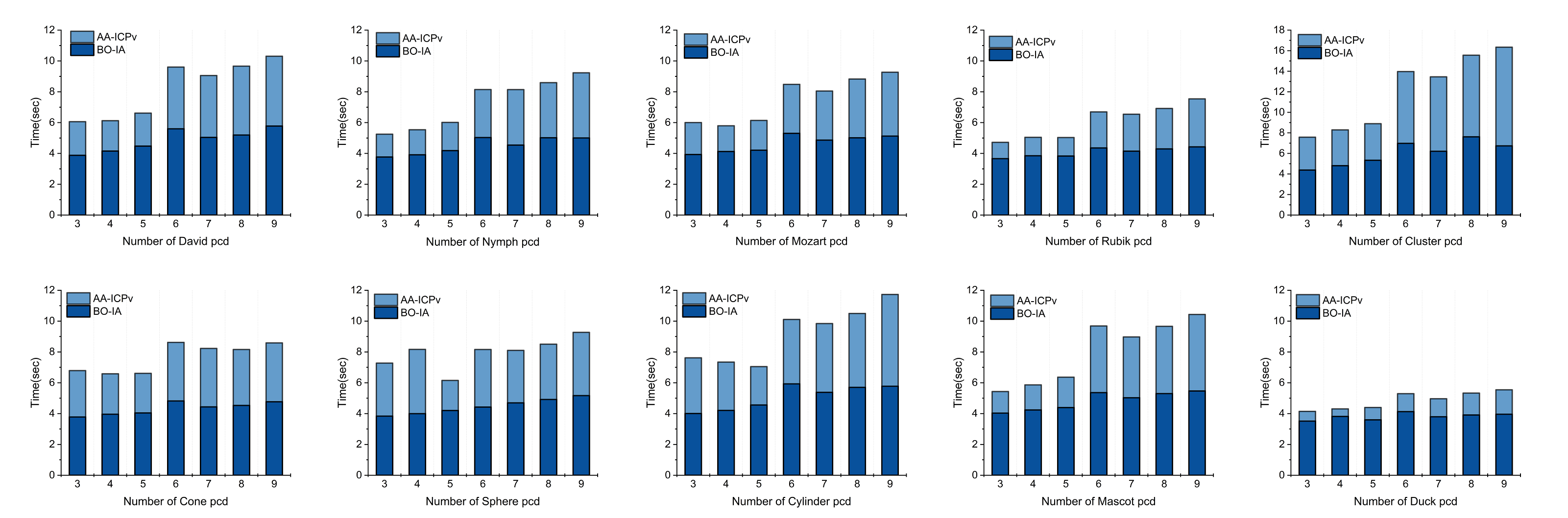}
	\caption{Median runtime of RegHEC and its breakdown when different scene and number of input point clouds.}
	\label{fig_13}
\end{figure*}

\begin{table*}[h!t]	
	\renewcommand\arraystretch{1.32}
	\small
	\centering
	\caption{Median/Average runtime of RegHEC when different scene and different number of input point clouds(unit:{\upshape sec})}
	\label{tab4}
	\resizebox{1.85\columnwidth}{!}{      
		\begin{tabular}{c c c c c c c c c }\hline
			\multirow{2}*{Object}& \multirow{2}*{Average pcd size}& \multicolumn{7}{c}{Numder of input point clouds} \\
			\cline{3-9} 
			&\qquad&3&4&5 &6 &7 &8 &9 \\ \hline
			David\qquad&17873&6.061/6.659&6.126/6.422&6.619/6.870&9.604/10.375&9.053/9.664&9.666/10.078&10.307/11.149\\
			Nymph\qquad&14643.9&5.249/5.411&5.537/5.668&6.013/6.251&8.144/8.430&8.141/8.571&8.592/9.306&9.232/10.096\\	
			Mozart\qquad&16211.2&6.005/6.163&5.796/6.235&6.143/6.338&8.482/8.712&8.050/8.577&8.828/9.072&9.267/10.087\\
			Rubik\qquad&9977.9&4.719/4.944&5.045/5.527&5.032/5.430&6.697/8.129&6.545/7.255&6.915/8.161&7.542/8.948\\		
			Cluster\qquad&26544.8&7.586/10.055&8.292/10.061&8.895/12.046&13.97/16.753&13.456/20.177&15.557/22.227&16.339/20.544\\
			Cone\qquad&11554.8&6.791/9.269&6.579/7.069&6.611/6.734&8.62/8.983&8.227/8.897&8.158/8.651&8.587/9.404\\
			Sphere\qquad&13581&7.276/8.11&8.164/9.540&6.152/6.920&8.158/9.553&8.099/11.465&8.503/20.693&9.271/12.133\\
			Cylinder\qquad&17266.8&7.62/13.090&7.342/7.938&7.047/8.038&10.106/11.428&9.842/11.674&10.497/11.657&11.727/15.284\\
			Mascot\qquad&16974.3&5.429/5.600&5.863/6.183&6.364/6.868&9.684/10.616&8.973/10.641&9.661/11.165&10.432/12.618\\
			Duck\qquad&6014&4.148/4.540&4.307/5.268&4.399/4.592&5.293/5.974&4.964/8.081&5.333/6.387&5.544/6.596\\
            \hline
	\end{tabular}  }                 
\end{table*}

Intuitively, RegHEC can be computationally expensive, as it contains both simultaneous initial alignment and simultaneous fine registration of multi-view point clouds, so we recorded execution time of all the 7000 runs then plotted the median runtime and its breakdowns when different scene and different number of input point clouds in Fig.13. The median/average runtime were tabulated in table II.

As expected, the algorithm's runtime shows an increasing trend as number of input point clouds or average point cloud size increases, due to more correspondences to be processed during both BO-IA and AA-ICPv. One can also see that AA-ICPv contributes more to this runtime increase, while the time consumption of BO-IA is rather stable and less affected. Unlike AA-ICPv which iteratively transforms multi-view point clouds into robot base frame, finds correspondences then solves minimization problem with tens of thousands of residual blocks, the increase in number of point clouds and point cloud size only means more time consumption in sampling procedure for BO-IA, whereas considerable execution time comes from EI maximization and hyper parameter optimization,  which are not affected by number of point cloud and point cloud size. The maximum runtime 16.339/20.544 was reported when 9 largest cluster point clouds were used.

Besides, the three SORs shows obviously longer runtime when small number of input point clouds compared with other object in similar size.  As mentioned earlier, 1.286\% (90) among the 7000 runs failed to converge within 100 iterations during AA-ICPv, we find that 54 of them come from the SORs when only 3 point clouds were used. It takes AA-ICPv generally more iterations to finally converge, leading to this abnormal runtime increase.

\subsection{Eye-to-hand scenario}
\begin{figure}[!t]
	\centering
	\includegraphics[width=0.5\textwidth]{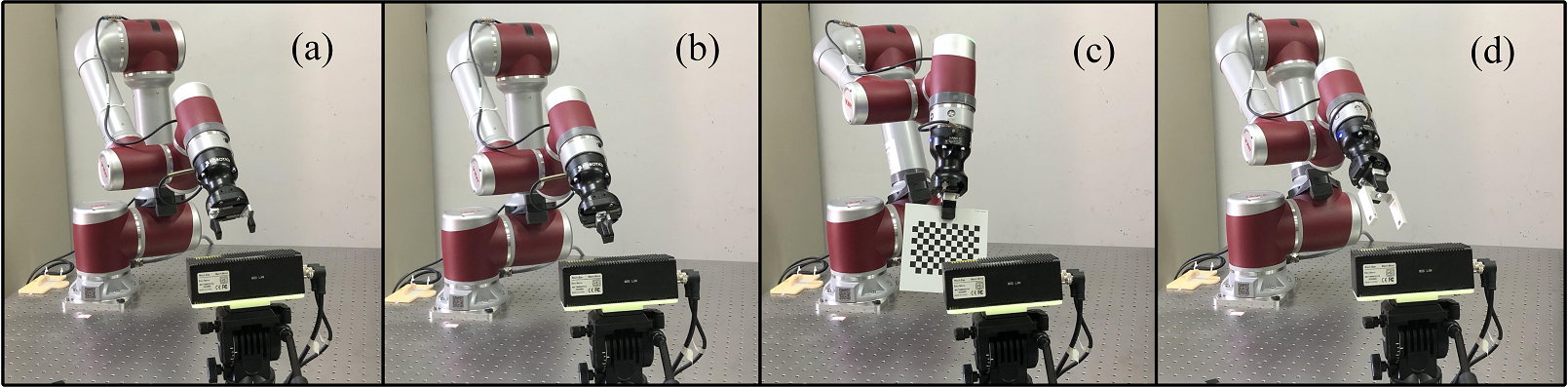}
	\caption{Experiments setup for eye-to-hand scenario. (a)Calibration with gripper open and RegHEC. (b)Calibration with gripper closed and RegHEC. (c)Calibration with checkerboard and other methods for comparison. (d)Data acquisition for accuracy assessment.}
	\label{fig_14}
\end{figure}

Although it is doable to carry out eye-to-hand experiment with arbitrary object in hand, we chose to use empty hand(gripper only), for two reasons. First, by assuming CAD model of gripper unavailable, we can directly take gripper as arbitrary reconstruction target, which is basically no different for RegHEC than arbitrary object in hand. Second, the combination of stationary 3-D sensor and end effector mounted on robot flange is a very common use case\cite{CircleDrawing,ContactMap}, thus it's meaningful to test RegHEC's performance in such scenario. 

The robot was controlled to place the gripper into 9 different poses in front of the stationary sensor and a point cloud was captured at each gripper poses.
We carried out above process with both gripper open and closed, as is shown in Fig.14(a) and (b). Like eye-in-hand experiment, point clouds were downsampled with voxel grid filter with leaf size of 0.001m. Points too far from sensor were filtered out as they are not always stationary w.r.t. robot flange when different robot poses.

\begin{table*}[h!t]	
	\renewcommand\arraystretch{1.32}
	\small
	\centering
	\caption{{Accuracy assessment ${Err}_{\boldsymbol R}$/${Err}_{\boldsymbol t}$ in eye-to-hand scenario. (unit:{$^\circ$/\upshape mm})}}
	\label{tab4}
	\resizebox{1.85\columnwidth}{!}{      
		\begin{tabular}{c c c c c c c c c }\hline
			\multirow{2}*{Method}& \multirow{2}*{Objects}& \multicolumn{7}{c}{Numder of input point clouds/images} \\
			\cline{3-9} 
			&\qquad&3&4&5 &6 &7 &8 &9 \\ \hline
			RegHEC&Gripper open\qquad&0.346/0.863 &0.170/0.499 &0.179/0.498 &0.178/0.473 &0.175/0.390 &0.181/0.363 &0.172/0.393\\
			RegHEC&Gripper closed\qquad&0.304/0.746 &0.209/0.388 &0.194/0.427 &0.176/0.425 &0.189/0.366 &0.198/0.338 &0.172/0.368\\
			Tsai\cite{Tsai}&Calibration board\qquad&0.261/1.626 &0.213/1.364 &0.166/0.985 &0.168/0.890 &0.167/0.852 &0.158/0.770 &0.156/0.754\\
			Horaud\cite{HoraudSimulIterative}&Calibration board\qquad&0.273/1.181 &0.237/1.202 &0.213/0.902 &0.183/0.613 &0.183/0.613 &0.156/0.560 &0.158/0.674\\
			Shah\cite{shah2013AXYBseparate}&Calibration board\qquad&0.271/1.449 &0.237/1.120 &0.191/0.792 &0.196/0.776 &0.196/0.764 &0.167/0.507 &0.164/0.491\\
			Li\cite{li2010AXYBsimul}&Calibration board\qquad&0.252/1.843 &0.228/1.054 &0.193/0.881 &0.196/0.841 &0.196/0.853 &0.167/0.644 &0.165/0.643\\
			Tabb\cite{tabb2017AXYBreproj}&Calibration board\qquad&0.196/0.928 &0.139/0.979 &0.142/0.725 &0.139/0.804 &0.139/0.802 &0.139/0.850 &0.142/0.737\\
			J.Wu\cite{wu2019handeyeSO4}&Calibration board\qquad&0.322/1.043 &0.285/1.298 &0.271/1.273 &0.235/0.934 &0.235/0.932 &0.178/0.560 &0.167/0.498\\
			L.Wu\cite{wu2016handeyeTmech}&Calibration board\qquad&0.255/1.256 &0.231/0.945 &0.189/0.870 &0.192/0.685 &0.193/0.683 &0.166/0.995 &0.163/1.031 \\
			
			\hline
	\end{tabular}  }                 
\end{table*}

\begin{table*}[h!t]	
	\renewcommand\arraystretch{1.32}
	\small
	\centering
	\caption{Median/Average runtime of RegHEC when different scene and number of input point clouds(unit:{\upshape sec})}
	\label{tab4}
	\resizebox{1.85\columnwidth}{!}{      
		\begin{tabular}{c c c c c c c c c }\hline
			\multirow{2}*{Object}& \multirow{2}*{Average pcd size}& \multicolumn{7}{c}{Numder of input point clouds} \\
			\cline{3-9} 
			&\qquad&3&4&5 &6 &7 &8 &9 \\ \hline
			Gripper open\qquad&17049.2&6.203/6.889&7.005/9.548&7.367/7.700&8.885/9.013&8.201/8.573&8.809/9.095&9.333/9.400\\
			Gripper closed\qquad&19349.6&6.207/6.978&7.339/8.655&7.234/7.746&9.442/9.649&8.707/9.187&9.122/9.446&9.891/10.276\\	
			\hline
	\end{tabular}  }                 
\end{table*}

For each number of point clouds(from 3 to maximum 9) of both gripper open and closed, we ran RegHEC 100 times using the same laptop with all parameters inherited from eye-in-hand experiment except the search space $\mathrm{b}$ during BO-IA was set to be $[-\pi,\pi]^3\times[-0.8,-0.4]\times[-0.4,0]\times[0,0.4]$. Please notice that, in eye-to-hand scenario bounds $\mathrm{b}$ roughly encloses sensor position w.r.t robot base frame and the inverse of corresponding robot poses were plugged in, as registration is achieved under robot flange frame.

Among the total 1400 runs, 99.143\% of them converged within 100 iterations and Fig.15 shows an example of registration result using RegHEC and all 9 point clouds.
\begin{figure}[!t]
	\centering
	\includegraphics[width=0.42\textwidth]{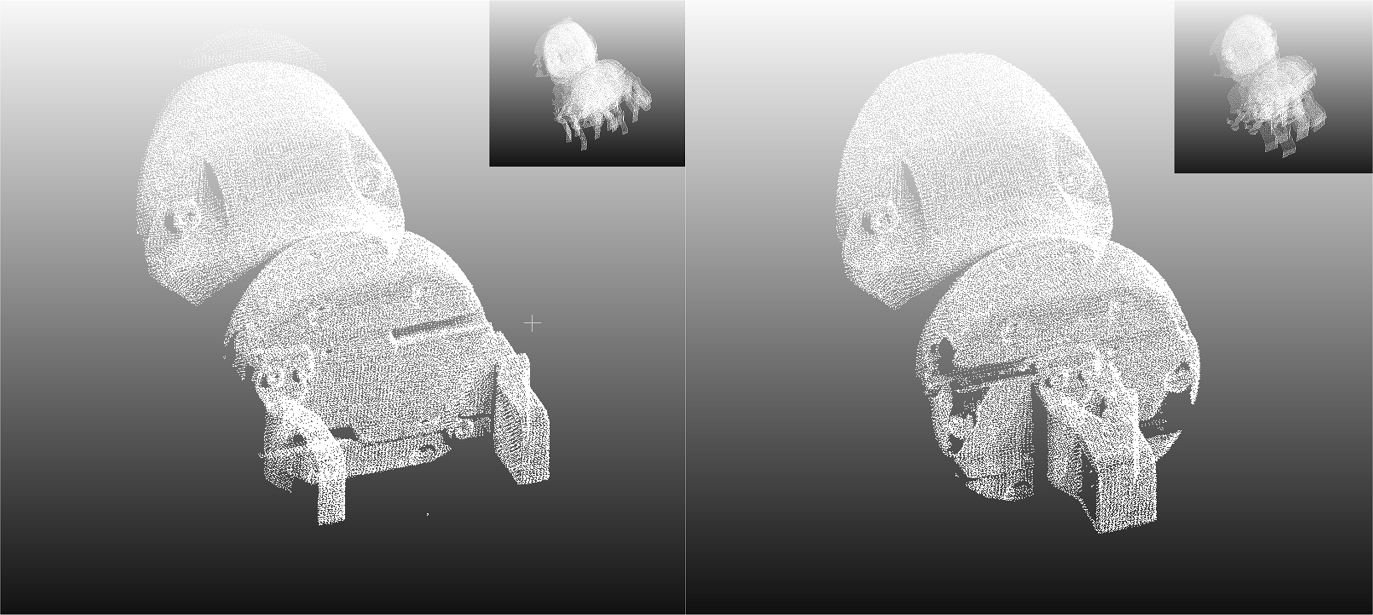}
	\caption{Registration of 9 point clouds from different viewpoints under robot flange frame, with RegHEC. Miniature in the top right shows the 9 source point clouds under sensor frame before registration.}
	\label{fig_15}
\end{figure} 
\begin{figure}[!t]
	\centering
	\includegraphics[width=0.45\textwidth]{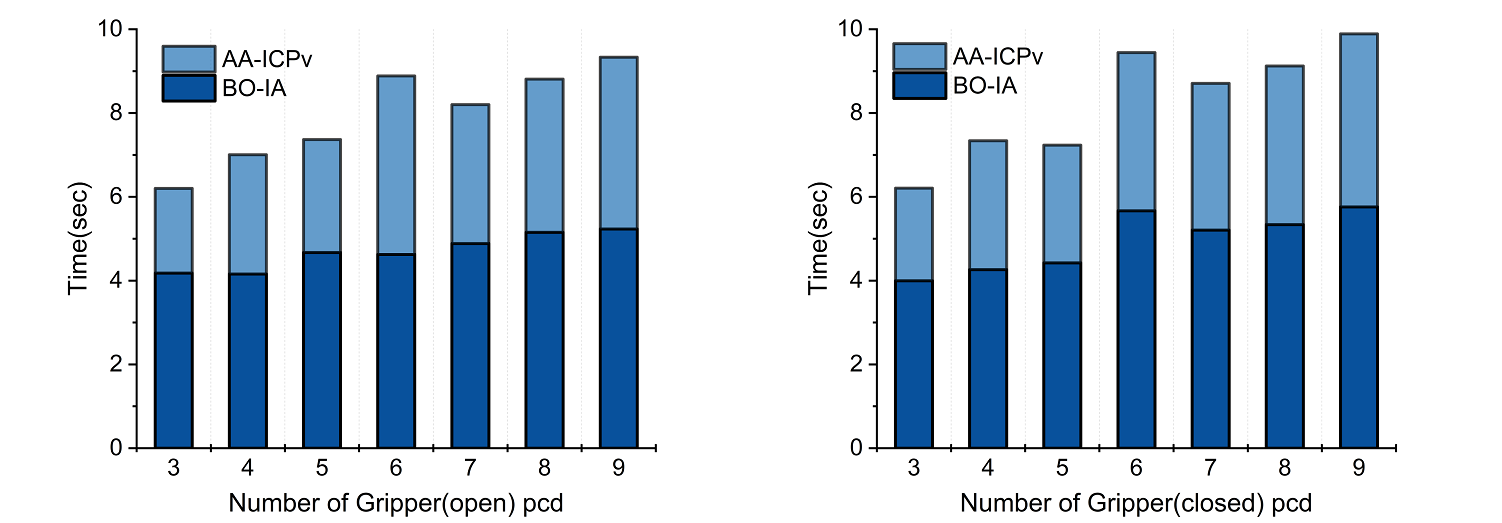}
	\caption{Median runtime of RegHEC and its breakdown when different scene and number of input point clouds.}
	\label{fig_16}
\end{figure}
Accuracy was still assessed from robotic 3-D reconstruction perspective and the same standard mechanical target was used again, as shown in Fig.14(d). This time, it was gripped firmly by the gripper and placed in front of the stationary sensor at 50 random robot poses different than those 9 for calibration. 50 Target point clouds were then transformed to robot flange frame using inverse of corresponding robot poses and base-sensor relation to be assessed. With sampled CAD model, we measured the 50 target poses w.r.t robot flange frame and took the SD of measured rotations and position as error metric, given in (19) and (20). All the 1400 calibrated base-sensor relation went through the assessment and results(median) were tabulated in Table III. Additionally, the 12*9 calibration checkerboard was used again to carry out calibrations with other 7 methods, as is shown in Fig.14(c). The checkerboard-based calibration results were also given in Table III for a clear comparison. In general, RegHEC shows similar accuracy compared with other checkerboard-based calibrations. Meanwhile, calibration performance still presents an increasing trend as number of input point clouds increases.
 
Execution time of all 1400 runs were recorded. The median/average runtime of each 100 runs were tabulated in Table IV. Fig.16 shows the breakdown of median time consumptions. Overall, RegHEC stopped within 10 seconds and the runtime increases as number of point clouds used or average pcd size increases. The maximum time cost is 9.891/10.276 seconds when 9 gripper closed point clouds with average size of 19349.6 were used.

Besides, it is worth noticing that the gripper point clouds were aligned into robot flange frame, which means that one can easily locate the tool center point(TCP) by selecting the TCP in any point cloud viewer. RegHEC doesn't only help the user to calibrate base-sensor transformation with no need for specialized calibration rig, but also makes TCP calibration easy when an end effector is installed and taken as reconstruction/calibration object.

\subsection{Performance under extreme condition}
Perhaps the most extreme and challenging scenario is a flat surface or plane, where correct point clouds registration and sensor motion estimation with existing pair-wise ICP seem impossible. However, as a very basic geometry, plane can be very common in daily life and robotic applications. For instance, an automated guided vehicle equipped with an eye-in-hand manipulator, standing by the wall and waiting for calibration, or an eye-in-hand manipulator installed on a neat workbench.

To test RegHEC's performance in such case, we prepared an A4 paper with handwriting "RegHEC" on it and point clouds were captured by the previous eye-in-hand setup from different viewpoints, as shown in Fig.17(a). The average size of captured point clouds is 11746.6 and the handwriting region is a blank space with no points, thus it helps to identify if the final registration is correct or not.
For each number of point clouds, we carried out 100 runs of RegHEC using same parameters as subsection A and all calibration results went through assessment described in subsection B.
\begin{figure}
	\centering
	\includegraphics [width=0.46\textwidth]{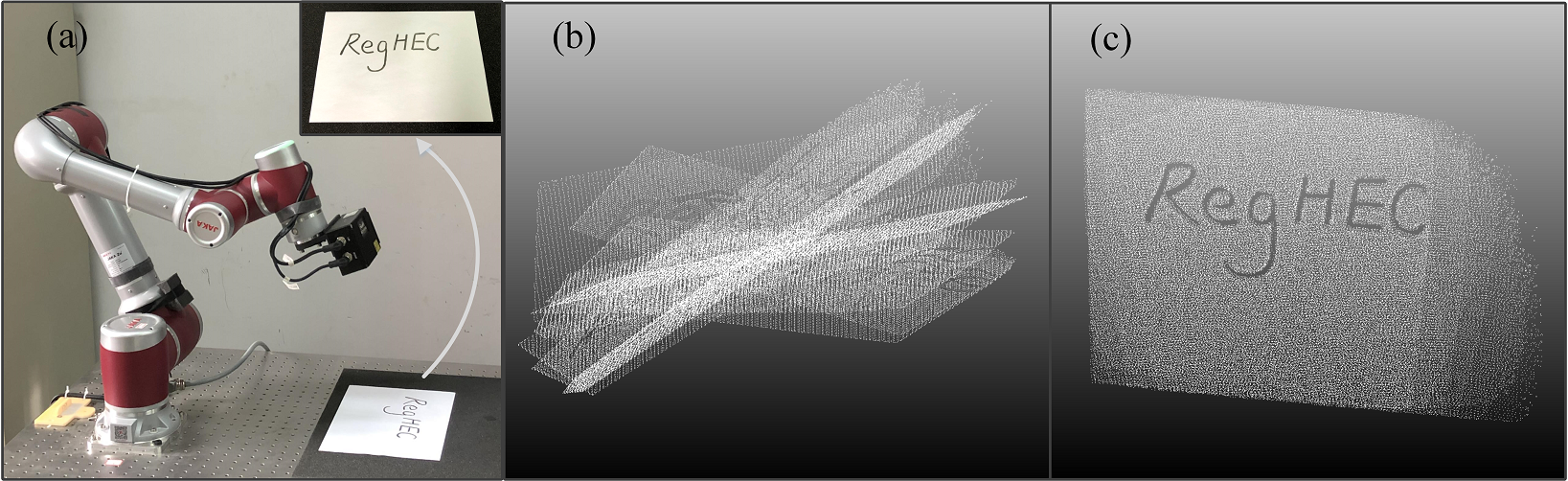}
	\caption{(a)An A4 paper with handwritten "RegHEC" is used to test our algorithm's performance facing flat surface or plane. (b)9 captured plane point clouds under sensor frame. (c)An example of Registration result when RegHEC converged.}
	\label{}       % Give a unique label
\end{figure}

\begin{figure}
	\centering
	\includegraphics [width=0.46\textwidth]{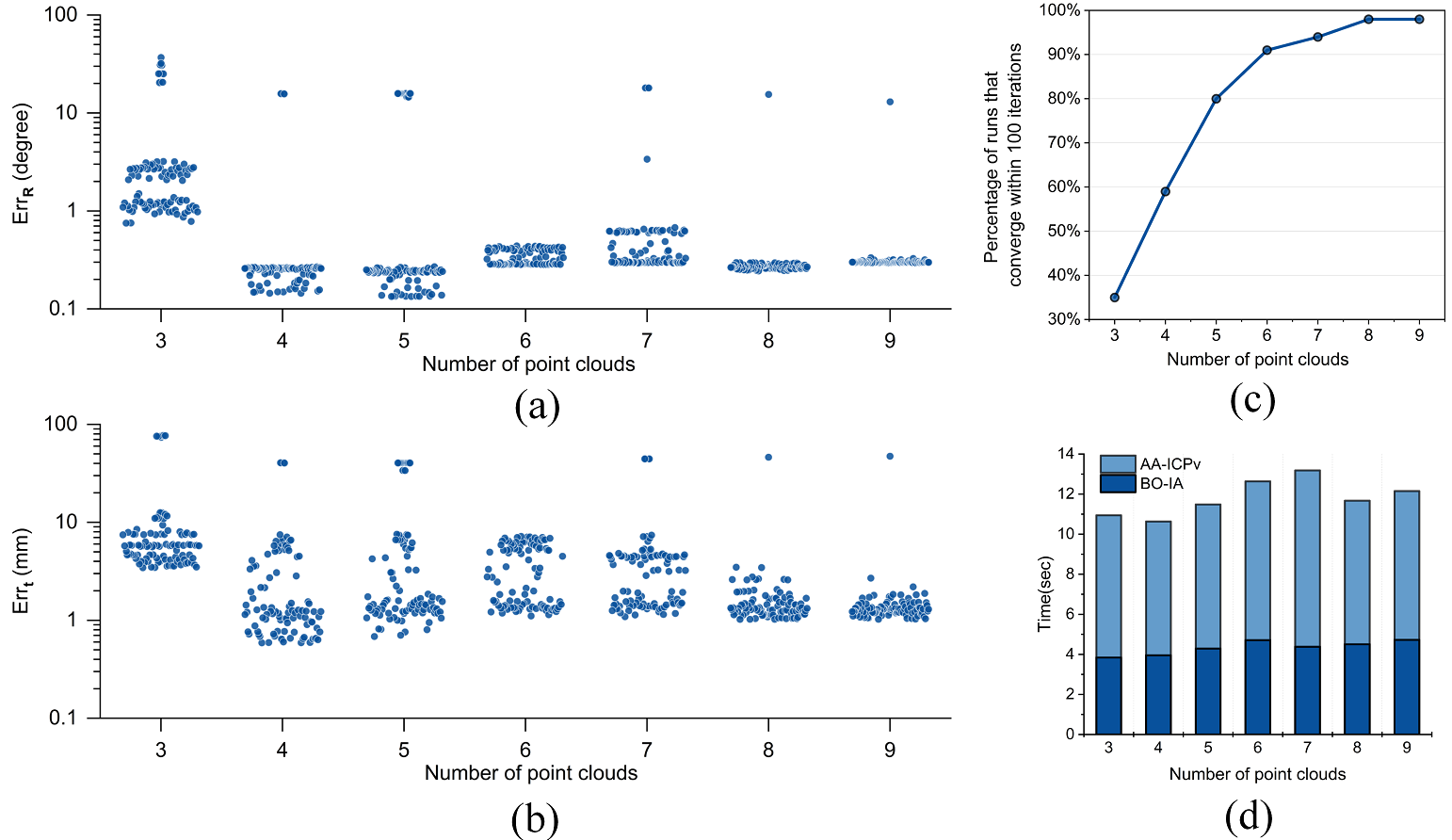}
	\caption{Results of all calibration runs. (a)${Err}_{\boldsymbol R}$ of all calibration runs. (b)${Err}_{\boldsymbol t}$ of all calibration runs. (c)Percentage of runs that converge within 100 AA-ICPv iterations. (d)Median runtime and breakdown.}
	\label{}       % Give a unique label
\end{figure}

Fig.17(b) and Fig.17(c) show the 9 source point clouds under sensor frame and an example of registration result with RegHEC when convergence respectively. The handwriting region is still clear, indicating 9 point clouds are correctly aligned. Although almost impossible for existing pair-wise ICP, plane scenario is less challenging and doable for RegHEC. 

Statistical properties of all calibration runs are given in Fig.18. Subplot (a) and (b) present ${Err}_{\boldsymbol R}$ and ${Err}_{\boldsymbol t}$ respectively, with y axis in log scale. Subplot(c) shows percentage of runs that converge within 100 iterations and subplot(d) shows the median runtime and breakdown. One can see that although only 35\% runs converged within 100 AA-ICPv iterations(when 3 input point clouds) which is even worse than the 3 SORs cases, the convergence got much easier as number of point clouds increased and 98\% runs converged within 100 iterations when 8 or 9 point clouds were used. This difficulty in convergence when small number of input point clouds is also reflected in longer runtime, as a result, the incresing trend in runtime is not so obvious as that of other tested objects. Generally, the calibration error exhibits a decreasing trend as number of point clouds increases, and the error in median when all 9 point clouds were used is $0.299^\circ/1.296\text{mm}$. Admittedly, RegHEC presents lower performance facing simple plane than other arbitrary objects, but is still much more promising than existing point cloud registration and hand-eye calibration algorithm under such extreme condition.

\section{Conclusions and future work}
RegHEC is a simultaneous multi-view point clouds registration and hand-eye calibration technique, applicable for both eye-in-hand and eye-to-hand cases. It only requires several point clouds of arbitrary object from different viewpoints and corresponding robot poses when capturing those point clouds. RegHEC helps to liberate our community from the specialized calibration rig, benefiting most of 3-D vision guided robotic applications, and is especially favorable for robotic 3-D reconstruction task, as it unifies the commonly separated calibration and reconstruction into a single process.

We hope RegHEC can bring more research opportunities and some directions for future improvement include but are not limited to:
\begin{enumerate}
	\item{Finding the closed form solution of (1) for faster convergence.}
	\item{Trying out different metric to establish optimization objective in (1) to enhance robustness, e.g. point-to-plane distance or $\mathcal{L}_0$ norm. }
	\item{Extension to more complicated multi-robot collaboration scenarios. Fig.19(a) demonstrates a collaboration scenario between a sensor robot and manipulation robot, where the calibration problem is usually formulated as AXB=YCZ for simultaneous hand–eye, tool–flange, and robot–robot calibration\cite{AXBYZC1,AXBYZC2}, and a calibration marker is a must in the existing work to define the tool frame. Fig.19(b) shows another collaborative 3-D reconstruction scenario, where 2 hand-eye relations and robot-robot relation are to be calibrated. RegHEC is a general idea which unifies point clouds registration and calibration, it is potentially feasible for above scenario with no need for additional calibration marker, after proper extension. The frame work of first initial alignment then fine registration still holds with more unknown transformations to be estimated.}
\end{enumerate}

\begin{figure}[!t]
	\centering
	\includegraphics[width=0.36\textwidth]{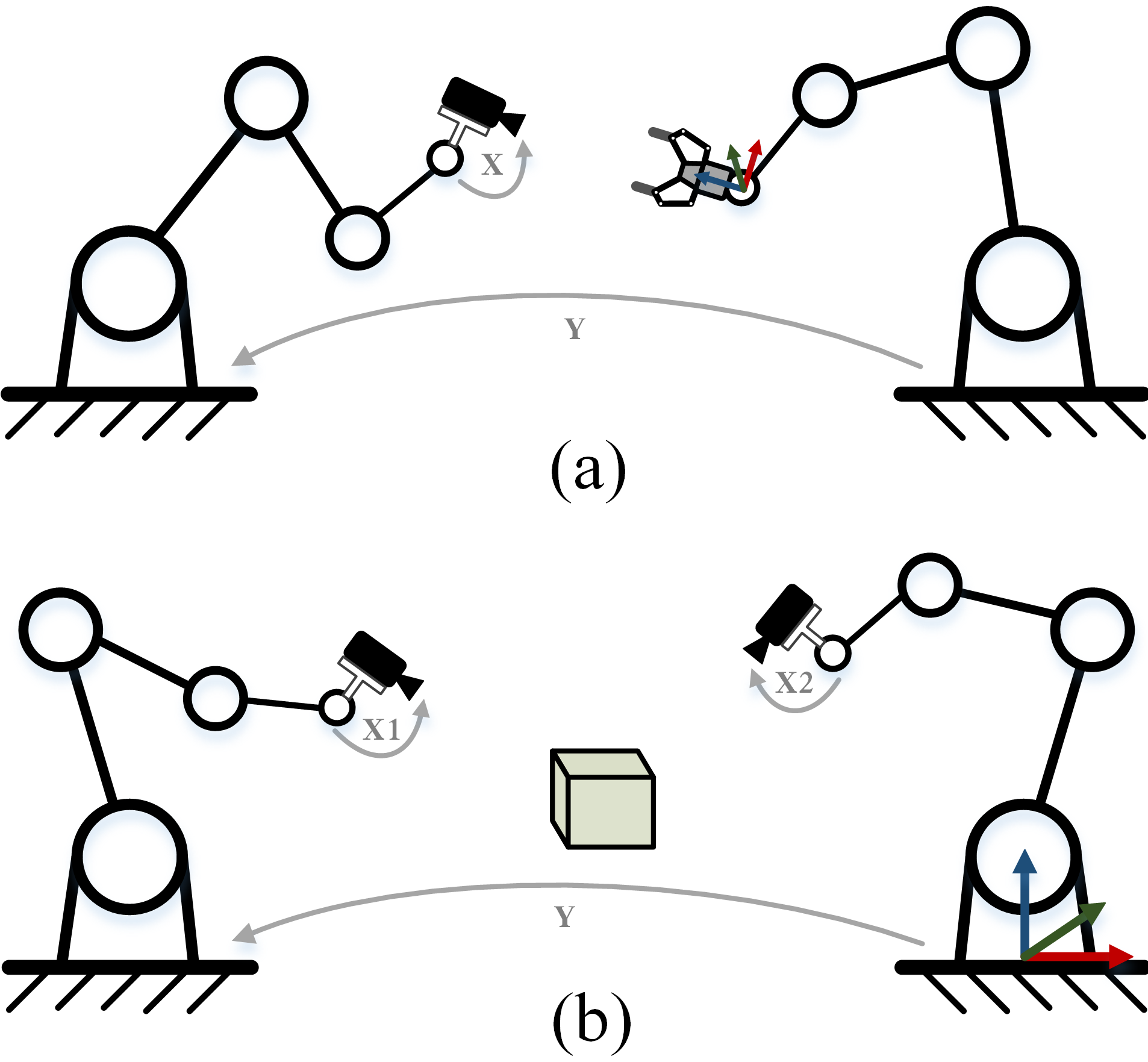}
	\caption{Potential use cases in multi-robot collaboration scenario. Registration reference frame is marked with colored arrow. (a)Use case with sensor robot and manipulation robot. Move the two manipulator into a set of poses, under which point cloud of end effector is taken by the 3-D sensor each time when both robots finish their motions, and then find the hand-eye X and robot-robot Y relation that is able to algin multi-view point clouds of end effector into the flange frame of manipulation robot. With registration of end effector point clouds, TCP can be calibrated in the meanwhile. (b)A collaborative 3-D reconstruction scenario. Likewise, find the two hand eye relations X1,X2 and robot-robot relation Y that align captured point clouds into robot base frame of either robot(we marked the base frame of robot on the right).}
	\label{fig_17}
\end{figure}

\section*{Acknowledgments}
The authors would like to thank Jean-Baptiste Mouret and Konstantinos Chatzilygeroudis for their help on kernel coding.

{\appendix[Solution to the least square problem in (1)]
In this appendix, we present a thorough derivation of perturbation-based Gauss-Newton method for solving least square problem (1).
	
First recall that each residual block is in form of
\begin{equation}
	\\[1ex]	
	{\begin{small}	
			\begin{split}	
				\begin{aligned}
					\!\!&\boldsymbol g_{ij}(\boldsymbol R_X \!,\!\boldsymbol t_X)\\[1ex]
					\!\!&= 
					\begin{bmatrix}
						% 	\begin{smallmatrix}
						\boldsymbol R_{A_i}\!\!& \!\!\boldsymbol t_{A_i}\\ 
						%	\end{smallmatrix}
					\end{bmatrix}
					\begin{bmatrix}
						%	\begin{smallmatrix}
						\boldsymbol R_{X}\!\!& \!\!\boldsymbol t_{X}\\ 
						\boldsymbol 0\!\!&\!\! 1 
						%	\end{smallmatrix}
					\end{bmatrix}
					\begin{bmatrix}
						% 	\begin{smallmatrix}
						\boldsymbol p_j^i\\ 
						1 
						%	\end{smallmatrix}
					\end{bmatrix}
					\!-\!\begin{bmatrix}
						%	\begin{smallmatrix}
						\boldsymbol R_{A_{i+1}}\!\!&\!\! \boldsymbol t_{A_{i+1}}\\ 
						%	\end{smallmatrix}
					\end{bmatrix}
					\begin{bmatrix}
						%	\begin{smallmatrix}
						\boldsymbol R_{X}\!\!&\!\! \boldsymbol t_{X}\\ 
						\boldsymbol 0\!\!&\!\! 1 
						%	\end{smallmatrix}
					\end{bmatrix}
					\begin{bmatrix}
						% 	\begin{smallmatrix}
						\boldsymbol q_j^i\\ 
						1 
						%	\end{smallmatrix}
					\end{bmatrix}\\[1ex]
					\!\!&=\boldsymbol R_{A_i}\boldsymbol R_{X}\boldsymbol p_j^i \!+\!\boldsymbol R_{A_i}\boldsymbol t_{X}\!+\!\boldsymbol t_{A_i}\!-\!\boldsymbol R_{A_{i+1}}\boldsymbol R_{X}\boldsymbol q_j^i \!-\!\boldsymbol R_{A_{i+1}}\boldsymbol t_{X}\!-\!\boldsymbol t_{A_{i+1}}
				\end{aligned}
			\end{split}	
	\end{small}}
\end{equation} 

Directly computing derivatives of above function with respective to $\boldsymbol R_{X}$ leads to a tricky optimization problem with 6 equality constraints. Thus, we consider $\boldsymbol R_{X}$ as constant and applied a small perturbation  $\boldsymbol {\varphi}_{X}$ in Lie Algebra so(3) on the left side of $\boldsymbol R_{X}$.  

\begin{equation}
	\\[1ex]	
	{\begin{small}	
			\begin{split}	
				\begin{aligned}
					\!\! \widetilde {\boldsymbol g}_{ij}(\boldsymbol {\varphi}_{X} ,\boldsymbol t_X)
					&= \boldsymbol R_{A_i}exp(\boldsymbol {\varphi}_{X}^\wedge)\boldsymbol R_{X}\boldsymbol p_j^i \!+\!\boldsymbol R_{A_i}\boldsymbol t_{X}\!+\!\boldsymbol t_{A_i}\\&-\boldsymbol R_{A_{i+1}}exp(\boldsymbol {\varphi}_{X}^\wedge)\boldsymbol R_{X}\boldsymbol q_j^i \!-\!\boldsymbol R_{A_{i+1}}\boldsymbol t_{X}\!-\!\boldsymbol t_{A_{i+1}}
				\end{aligned}
			\end{split}	
	\end{small}}
\end{equation} 
where ${(\cdot)}^\wedge$ means skew symmetric matrix and $exp(\cdot^\wedge)$ is exponential mapping from so(3) to SO(3)\cite{sola2018microLie}. Now we have an objective function over 6-dimensional vector with perturbation in Lie algebra as its head and translation vector of hand-eye relation as its tail, without any constraints.

Take partial derivative with respect to $\boldsymbol {\varphi}_{X}$ at zero perturbation, we have 
\begin{equation}
	\begin{split}	
		\begin{aligned}
 \frac{\partial \widetilde{\boldsymbol g}_{ij}}{\partial \boldsymbol {\varphi}_{X}}\Big|_ {\boldsymbol \varphi=\boldsymbol 0}&=\lim\limits_{\delta \boldsymbol \varphi_{X} \to \boldsymbol 0} \frac{\widetilde{\boldsymbol g}_{ij}(\boldsymbol 0+\delta \boldsymbol \varphi_{X},\boldsymbol t_{X})-\widetilde{\boldsymbol g}_{ij}(\boldsymbol 0,\boldsymbol t_{X})}{\delta \boldsymbol \varphi_{X}}\\
 &=\frac{\parbox{6cm}{$\boldsymbol R_{A_i}exp(\delta \boldsymbol \varphi_{X}^\wedge)\boldsymbol R_{X}\boldsymbol p_j^i-\boldsymbol R_{A_i}\boldsymbol R_{X}\boldsymbol p_j^i-\boldsymbol R_{A_{i+1}}exp(\delta \boldsymbol \varphi_{X}^\wedge)\boldsymbol R_{X}\boldsymbol q_j^i+\boldsymbol R_{A_{i+1}}\boldsymbol R_{X}\boldsymbol q_j^i$}}{\delta \boldsymbol \varphi_{X}}\\
 &=\frac{\parbox{6cm}{$\boldsymbol R_{A_i}(\boldsymbol I + {\delta \boldsymbol \varphi_{X}}^\wedge)\boldsymbol R_{X}\boldsymbol p_j^i-\boldsymbol R_{A_i}\boldsymbol R_{X}\boldsymbol p_j^i-\boldsymbol R_{A_{i+1}}(\boldsymbol I + {\delta \boldsymbol \varphi_{X}}^\wedge)\boldsymbol R_{X}\boldsymbol q_j^i+\boldsymbol R_{A_{i+1}}\boldsymbol R_{X}\boldsymbol q_j^i$}}{\delta \boldsymbol \varphi_{X}}\\
 &=\frac{\parbox{6.5cm}{$\boldsymbol R_{A_i}({\delta \boldsymbol \varphi_{X}}^\wedge)\boldsymbol R_{X}\boldsymbol p_j^i-\boldsymbol R_{A_{i+1}}({\delta \boldsymbol \varphi_{X}}^\wedge)\boldsymbol R_{X}\boldsymbol q_j^i$}}{\delta \boldsymbol \varphi_{X}}\\
 &=\frac{\parbox{6.5cm}{$-\boldsymbol R_{A_i}({\boldsymbol R_{X}\boldsymbol p_j^i})^\wedge\delta \boldsymbol \varphi_{X}+\boldsymbol R_{A_{i+1}}({\boldsymbol R_{X}\boldsymbol q_j^i})^\wedge\delta \boldsymbol \varphi_{X}$}}{\delta \boldsymbol \varphi_{X}}\\
 &=	-\boldsymbol R_{A_i}{(\boldsymbol R_{X}\boldsymbol p_j^i)}^\wedge +	\boldsymbol R_{A_{i+1}}{(\boldsymbol R_{X}\boldsymbol q_j^i)}^\wedge 
 	\end{aligned}
\end{split}	
\end{equation}
Take partial derivative with respect to $\boldsymbol t_{X}$, we have

\begin{equation}
	\frac{\partial \widetilde{\boldsymbol g}_{ij}}{\partial \boldsymbol t_X}=
	\begin{bmatrix}
    \boldsymbol R_{A_i} - \boldsymbol R_{A_{i+1}}
	\end{bmatrix}
\end{equation}
The Jacobian at $\begin{bmatrix}
	% 	\begin{smallmatrix}
	\boldsymbol 0\!\!& \!\!\boldsymbol t_{X}\\ 
	%	\end{smallmatrix}
\end{bmatrix}$ is then
\begin{equation}
	{\begin{footnotesize}	
			\!\!\! \boldsymbol J_{ij}(\boldsymbol \varphi_X ,\!\boldsymbol t_X)\!=\!				
			\begin{bmatrix}
				-\!\boldsymbol R_{A_i}{(\boldsymbol R_{X}\boldsymbol p_j^i)}^\wedge \!\!+\!\!	\boldsymbol R_{A_{i+1}}{(\boldsymbol R_{X}\boldsymbol q_j^i)}^\wedge \!& \!\boldsymbol R_{A_i}\!\!-\!\!\boldsymbol R_{A_{i+1}}
			\end{bmatrix}
	\end{footnotesize}}
\end{equation}
and the normal equation reads
\begin{equation}
	\begin{small}	
		(\sum_{i=1}^n\sum_{j=1}^{m_i}{\boldsymbol J_{ij}}^\mathrm{T}\boldsymbol J_{ij})\Delta\boldsymbol{X}=-\sum_{i=1}^n\sum_{j=1}^{m_i}{\boldsymbol J_{ij}}^\mathrm{T}  \boldsymbol g_{ij}
	\end{small}	
\end{equation}

With obtained next best step $\Delta\boldsymbol{X}$ in form of ${\begin{bmatrix}
		\Delta\boldsymbol{\varphi}^\mathrm{T}&{\Delta\boldsymbol{t}}^\mathrm{T}
\end{bmatrix}}^\mathrm{T}$, the rotation component and translation component of hand eye relation is updated into $exp({\!\Delta\boldsymbol{\varphi}}^\wedge)\boldsymbol R_X$ and $\boldsymbol t_X \!\!\!+\!\!\! \Delta\boldsymbol{t}$ respectively. Then go back to (22) and repeat until $\Delta\boldsymbol{X}$ is small enough. The algorithm decreases the target function in (1) monotonically till convergence and is summarized in Algorithm 3.
\begin{algorithm}
	\small
	\textbf{Input:}	$\{\boldsymbol {A_1, A_2,...,A_{n+1}}\}$ and $\{({\boldsymbol p_j^i},{\boldsymbol q_j^i})\}(i=1,2,...,n, n\geq2)$;\\Initial guess $(\boldsymbol R_X \!,\!\boldsymbol t_X)$ and stop threshold $\xi$\\
	\textbf{Output:} Optimized hand-eye relation	
	\caption{Optimize hand-eye relation $\boldsymbol{X}$ to align multiple point sets with known correspondences under robot base frame}             	
	\label{algorithm 1: }                        		
	\begin{algorithmic}[1]		
		\WHILE{ $||\Delta \boldsymbol X|| \ge \xi$ }
		
		\STATE \quad\quad  $\boldsymbol g_{ij} \longleftarrow \boldsymbol A_i, \boldsymbol A_{i+1}, (\boldsymbol p_j^i,\boldsymbol q_j^i), \boldsymbol R_X, \boldsymbol t_X$

		\STATE \quad\quad$\boldsymbol J_{ij} \longleftarrow \boldsymbol A_i, \boldsymbol A_{i+1}, (\boldsymbol p_j^i,\boldsymbol q_j^i), \boldsymbol R_X$
		
		\STATE \quad\quad$\Delta \boldsymbol X = -{(\sum_{i=1}^n\sum_{j=1}^{m_i}{\boldsymbol J_{ij}}^\mathrm{T}\boldsymbol J_{ij})}^{-1}\sum_{i=1}^n\sum_{j=1}^{m_i}{\boldsymbol J_{ij}}^\mathrm{T}  \boldsymbol g_{ij}$
		
		\STATE \quad\quad Update:
		\STATE \quad\quad$\boldsymbol R_X \longleftarrow exp({\Delta \boldsymbol{\varphi}}^{\wedge})\boldsymbol R_X$
		\STATE \quad\quad$\boldsymbol t_X \longleftarrow \boldsymbol t_X + \Delta \boldsymbol t$
		\ENDWHILE
		\RETURN $\boldsymbol R_X , \boldsymbol t_X$
		
	\end{algorithmic}	
\end{algorithm}

It is worth noticing that the derivatives with respect to $\boldsymbol{\varphi}$ in (23) holds at very small perturbation, thus we update the component residual function in (22) with the updated $\boldsymbol R_X$ and always take first order Taylor expansion at zero perturbation to linearize the model, which is slightly different than the traditional Gauss-Newton method where model function stays unchanged and iterate is updated in every iteration.

\bibliographystyle{IEEEtran}
\bibliography{IEEEabrv,RegHEC}

%\newpage

\vfill

\end{document}